\documentclass[10pt,twocolumn,letterpaper]{article}

\usepackage{cvpr}
\usepackage{times}
\usepackage{epsfig}
\usepackage{graphicx}
\usepackage{amsmath}
\usepackage{amssymb}

\usepackage{algorithm}
\usepackage{algorithmic}
\newtheorem{definition}{Definition}
\usepackage{color}

\usepackage[pagebackref=true,breaklinks=true,letterpaper=true,colorlinks,bookmarks=false]{hyperref}

\cvprfinalcopy 

\def\cvprPaperID{****} 

\ifcvprfinal\fi
\begin{document}

\title{Visual Graph Mining}

\author{Quanshi Zhang\dag\ddag, Xuan Song\dag, and Ryosuke Shibasaki\dag\\
\dag University of Tokyo, \qquad \ddag University of California, Los Angeles
}

\maketitle

\begin{abstract}
In this study, we formulate the concept of ``mining maximal-size frequent subgraphs'' in the challenging domain of visual data (images and videos). In general, visual knowledge can usually be modeled as attributed relational graphs (ARGs) with local attributes representing local parts and pairwise attributes describing the spatial relationship between parts. Thus, from a practical perspective, such mining of maximal-size subgraphs can be regarded as a general platform for discovering and modeling the common objects within cluttered and unlabeled visual data. Then, from a theoretical perspective, visual graph mining should encode and overcome the great fuzziness of messy data collected from complex real-world situations, which conflicts with the conventional theoretical basis of graph mining designed for tabular data. Common subgraphs hidden in these ARGs usually have soft attributes, with considerable inter-graph variation. More importantly, we should also discover the latent pattern space, including similarity metrics for the pattern and hidden node relations, during the mining process. In this study, we redefine the visual subgraph pattern that encodes all of these challenges in a general way, and propose an approximate but efficient solution to graph mining. We conduct five experiments to evaluate our method with different kinds of visual data, including videos and RGB/RGB-D images. These experiments demonstrate the generality of the proposed method.
\end{abstract}

\section{Introduction}
\label{sec:intro}

Graph mining is a classical field in data mining. To ease the mining process, pioneering techniques generally mined tabular data in which graphs contain distinct node and edge labels. However, the visual data collected from real-world situations, such as images and videos, are much fuzzier. This fuzziness undermines the basis of conventional mining approaches. Therefore, in this study, we propose a general formulation for the visual fuzziness, and extend the application scope of graph mining to visual data.

\begin{figure}
\centering
\includegraphics[width=\linewidth]{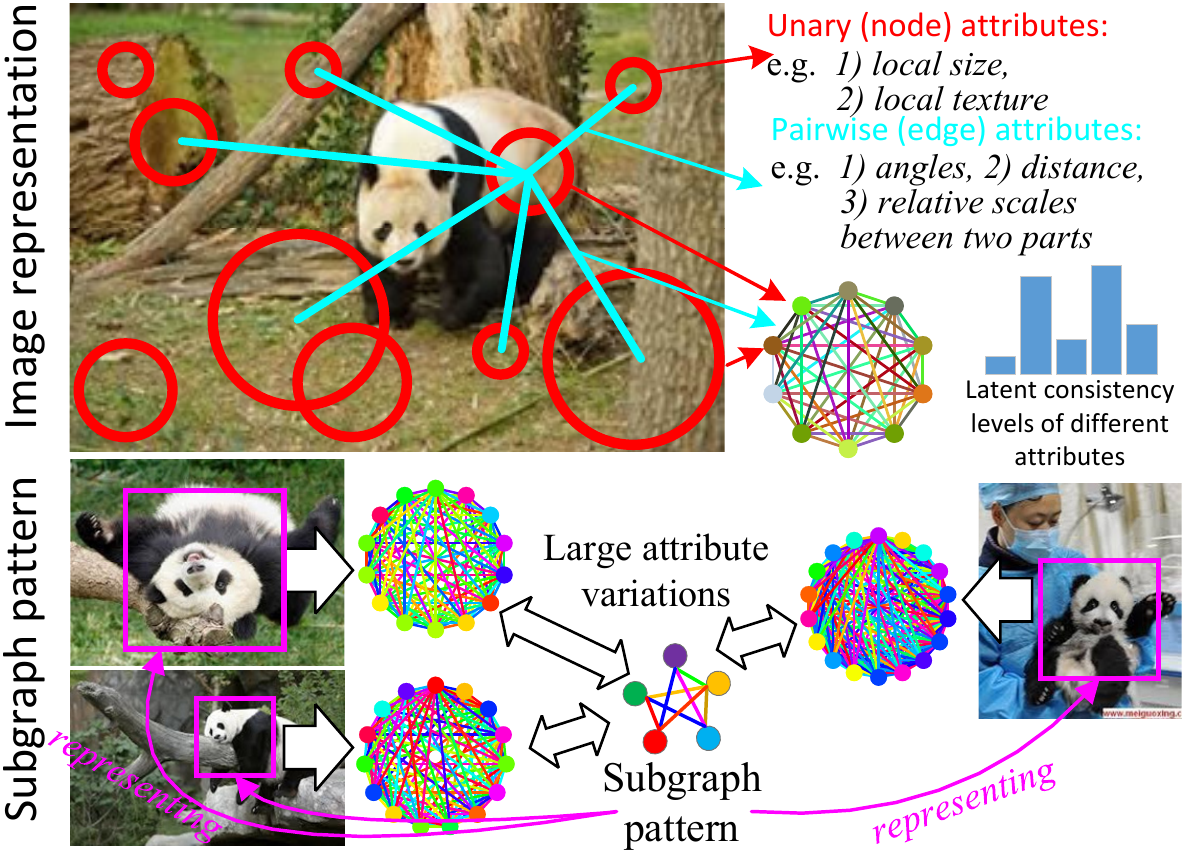}
\caption{Understanding the mVAP among ARGs. Node/edge colors denote different unary/pairwise attributes.}
\label{fig:top}
\end{figure}

\textbf{Concept of subgraph patterns:}{\verb| |} Before the introduction of the theory, let us first see a typical application, \emph{i.e.} mining detailed objects from unlabeled images. An image can be modeled as an attributed relational graph (ARG), as shown in Fig.~\ref{fig:top}. Each node in the ARG contains a number of high-dimensional unary attributes to describe different local features. Pairwise attributes on the edges measure different spatial relationships between object parts. In this case, the model for the common objects in the images corresponds to the subgraph pattern among the ARGs. In other words, ``mining frequent subgraph patterns among visual ARGs'' can be regarded as an elegant solution to ``mining and modeling objects with similar appearance and structure from unlabeled cluttered big visual data.''

Strictly speaking, mining \textit{detailed} object knowledge is a serious challenge, if the target objects are unlabeled and randomly located across large and cluttered images.

Unlike tabular data, visual data presents an intuitive problem, \emph{i.e.} we should simultaneously consider object occlusions and continuous changes in texture, rotation, scale, and pose among different objects. Such variations are formulated as attribute variations among the ARGs. Zhang \emph{et al.}~\cite{OurCVPR14Graph} have attempted to mine subgraphs with such variations.

However, in real-world situations, the bigger challenge lies in the uncertainty of pattern similarity metrics. Different subgraph patterns usually have their own metrics to evaluate the similarity between subgraphs. First, we should discover the hidden dependency/linkage relations between nodes. The selective use of strong part dependencies (\emph{e.g.} strong spatial relationships (or edges) between the head node and the body node) and neglect of weak linkages (\emph{e.g.} weak spatial relationships (or edges) between forefeet and hind feet) would produce stable mining performance. Second, we should incrementally discover latently effective attributes to guide the mining process. For example, the spatial relationship between parts may be the key factor identifying patterns of rigid objects, but is not so significant when measuring the pattern similarity between dynamic animals.

\textbf{Visual graph mining:}{\verb| |} The key to this research is the generality of our method, as this ensures its broad applicability. Therefore, in this study, we define a general fuzzy visual attributed pattern (VAP) to comprehensively formulate all the above visual challenges that are ubiquitous in different visual data.

In addition, we develop a general method to efficiently mine such patterns. Given an initial graph template, we gradually modify this template to the maximal-size VAP (mVAP) by discovering new nodes, eliminating redundant nodes, adjusting node linkages, and training attributes and matching parameters. This is different from previous methods for mining knowledge from visual data, which selectively modeled certain knowledge in a specific kind of visual data and neglected other variations (see Section~\ref{sec:related} for further discussion).

However, the mVAP's comprehensive modeling of visual challenges raises graph-mining difficulties to a new level. First, approaches to mining from tabular data usually require the nodes or edges to have distinct labels, and use such labels to enumerate new nodes for the pattern. However, this node enumeration strategy is hampered by visual fuzziness\footnote{This strategy uses local labels on nodes or edges to search new nodes for the pattern. However, in each specific visual ARG, both unary (node) attributes and pairwise (edge) attributes may be heavily biased. In addition, due to the existence of part (node) occlusion in visual ARGs, it is difficult to reduce the computational load by limiting the node enumeration within any single ARG.}.

\begin{figure}
\centering
\includegraphics[width=0.9\linewidth]{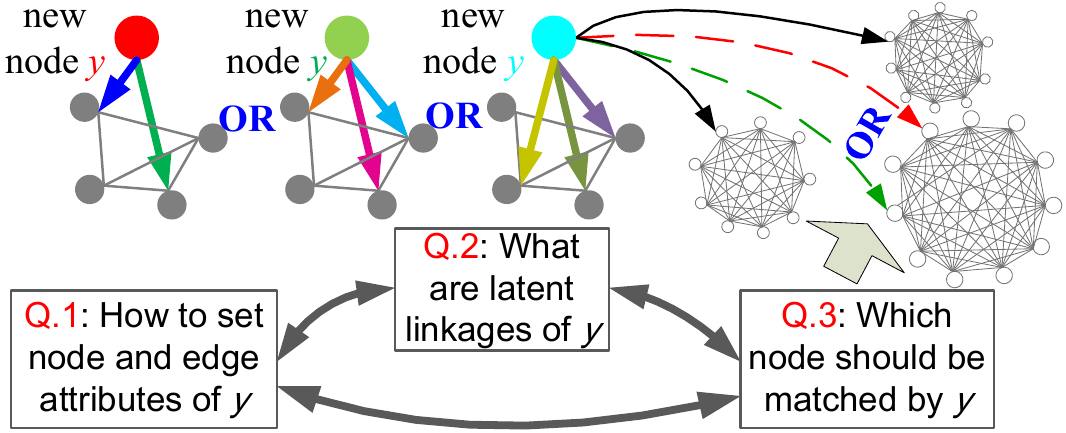}
\caption{Three-term chicken-and-egg problem in node discovery that raises the graph-mining challenge to a new level. We omit colors in unimportant nodes/edges for clarity.}
\label{fig:three}
\end{figure}

Second, visual patterns in pioneering studies, such as \cite{OurCVPR14Graph} and \cite{OurICCV13}, can be approximately described by Definition~\ref{def:VAP}(a), and the latent terms presented in Definition~\ref{def:VAP}(b,c) make this a \textbf{three-term} chicken-and-egg problem\footnote{1) Given pattern attributes and linkages, we can use graph matching techniques to compute the optimal matching assignments; 2) Given the matching assignments, the pattern attributes can be modified to better represent corresponding subgraphs in the ARGs; 3) Given the pattern attributes and matching assignments, strong linkages between pattern nodes can be determined. Please see the supplementary materials for details.} for discovering a new node $y$ for the pattern, as shown in Fig.~\ref{fig:three}. The three interdependent terms include 1) the determination of the matching assignments mapping $y$ to different ARGs, 2) the learning of the unary and pairwise attributes of $y$, and 3) the discovery of $y$'s latent linkages. Moreover, these three mutually influenced terms are hidden in great visual variations. Thus, we must simultaneously estimate them, which is NP hard. Fortunately, we have demonstrated an approximate but efficient solution to this problem that does not exhaustively enumerate pattern nodes and linkages.

\textbf{Summary:}{\verb| |} The contributions of this study can be summarized as follows. We formulate a novel subgraph pattern in a general form that has sufficient expressive power to comprehensively model the latent object knowledge in fuzzy visual data. The technical challenges of graph mining are significantly increased by real-world data fuzziness, and we propose an efficient approximate solution to the NP-hard mining problem. The generality of the proposed method is tested by different visual ARGs in different experiments.

The remainder of this paper is organized as follows. The next section discusses some related work. Section~\ref{sec:VAP} defines the subgraph pattern for visual data, and Section~\ref{sec:mining} presents the graph mining algorithm. We design five experiments to evaluate the proposed method in Section~\ref{sec:exper}. Finally, the overall study is summarized in Section~\ref{sec:conclude}.

\section{Related work}
\label{sec:related}

\textbf{Graph mining:} In the field of graph mining (reviewed in \cite{FrequentSubgraphMiningMain}), conventional ideas of mining maximal subgraph patterns have been realized as maximal frequent subgraph (MFS) extraction~\cite{MFSMiningMain,GraphMining1,GraphMining2,GraphMining4} and the mining of maximal cliques~\cite{CliqueMiningMain,CliqueMiningMain2}. MFS extraction approaches~\cite{MFSMiningMain,MFSMining1,GraphMining1} usually require the graphs to contain distinct node/edge labels or use local consistency to determine a set of node correspondence candidates between different graphs. Moreover, the graphs must have distinguishing structures. Thus, these methods define MFSs using graph isomorphisms. They mainly enumerate the nodes from different graphs to search the subgraphs with isomorphic (or similar) structures and labels. The distinct labels are used to prune the search range and avoid the NP-hard computation in the worst case. Similarly, the mining of maximal cliques ~\cite{CliqueMiningMain,CliqueMiningMain2,CliqueMining4,CliqueMining5} mainly extracts dense cliques that maintain geometric consistency. Subgraph patterns for ARGs have been defined~\cite{MiningARG1,MiningARG2}, and the softness of clique patterns has been formulated~\cite{SoftGraphMining1,SoftGraphMining2}.

However, the above methods are oriented to tabular data, and cannot be applied to fuzzy visual data\footnote{In this study, we focus on general visual data with great fuzziness, rather than the simplest visual data used in \cite{MFSMining1,MFSMining2}.} because of the requirement for node or edge labels or potential node correspondences. Visual ARGs may have considerable attribute variations, and cannot provide node correspondences in a local manner. Generally, node matching between ARGs is formulated as a quadratic assignment problem (QAP), which should be solved via a global optimization (as in (\ref{eqn:matching}), (\ref{eqn:new}), and (\ref{eqn:new_refine})). Thus, we must reformulate the whole theory on the basis of graph matching. In addition, as in \cite{OurCVPR14Graph}, the visual graph mining usually requires a rough graph template to start the whole mining process.

\textbf{Learning graph matching:} Given a graph template and a number of ARGs, methods for learning graph matching have been proposed to train parameters or refine the graph template for better matching performance. Most techniques~\cite{Cho1,LearnMatchingMaxMargin,LearnMatchingCMU,LearnMatchingNIO,LearningGraphMatchingSmooth} take a supervised approach, \emph{i.e.} they require the manual labeling of node correspondences between different ARGs. Leordeanu \emph{et al.}~\cite{LearnMatchingCMU} proposed the first unsupervised method of learning graph matching, and Zhang \emph{et al.}~\cite{OurICCV13} further refined the template structure in an unsupervised fashion. Cho \emph{et al.}~\cite{ProgressiveGraphMatching} proposed a similar idea that matched two ARGs and simultaneously extracted the most reliable edges between the two matched subgraphs. Essentially, these methods are not comparable with graph mining. They mainly train parameters or delete ``bad'' nodes from the graph template, rather than discovering new pattern nodes and recovering the prototype graphical patterns.

\textbf{Visual mining:} From the perspective of applications, there are numerous ways of mining objects from unlabeled big visual data, such as object discovery~\cite{i1}, co-segmentation~\cite{Coseg2}, edge model extraction~\cite{i12}, the learning of structural patterns~\cite{Beyond,LearnGraph}, and a number of techniques~\cite{CategoryModelingWithLink,CliqueMining1,CliqueMining2,CliqueMining3,MatchingRepetitive,LearnStructureFromImage} related to maximal clique mining. However, these studies were mainly designed with some specific techniques oriented to their own applications. They may model the structural knowledge by ignoring texture variations, or model the textural knowledge by ignoring structure deformation.

In contrast, graph mining has a clear expressive power in describing objects with exact shapes, which elegantly encodes all variations in texture, rotation, scale, and pose. The only pioneering work on graph mining~\cite{OurCVPR14Graph} simply assumed that common objects had no significant deformation between any pair of parts and did not discover hidden weights for attributes. Thus, its application was limited to simple rigid objects. However, in this study, we break this bottleneck of algorithm generality. We define a general subgraph pattern that formulates different kinds of visual fuzziness in a general form and develop a general method to mine such patterns. The generality of our method is demonstrated in five experiments.

\section{Maximal-size subgraph pattern}
\label{sec:VAP}

\begin{figure}
\centering
\includegraphics[width=0.9\linewidth]{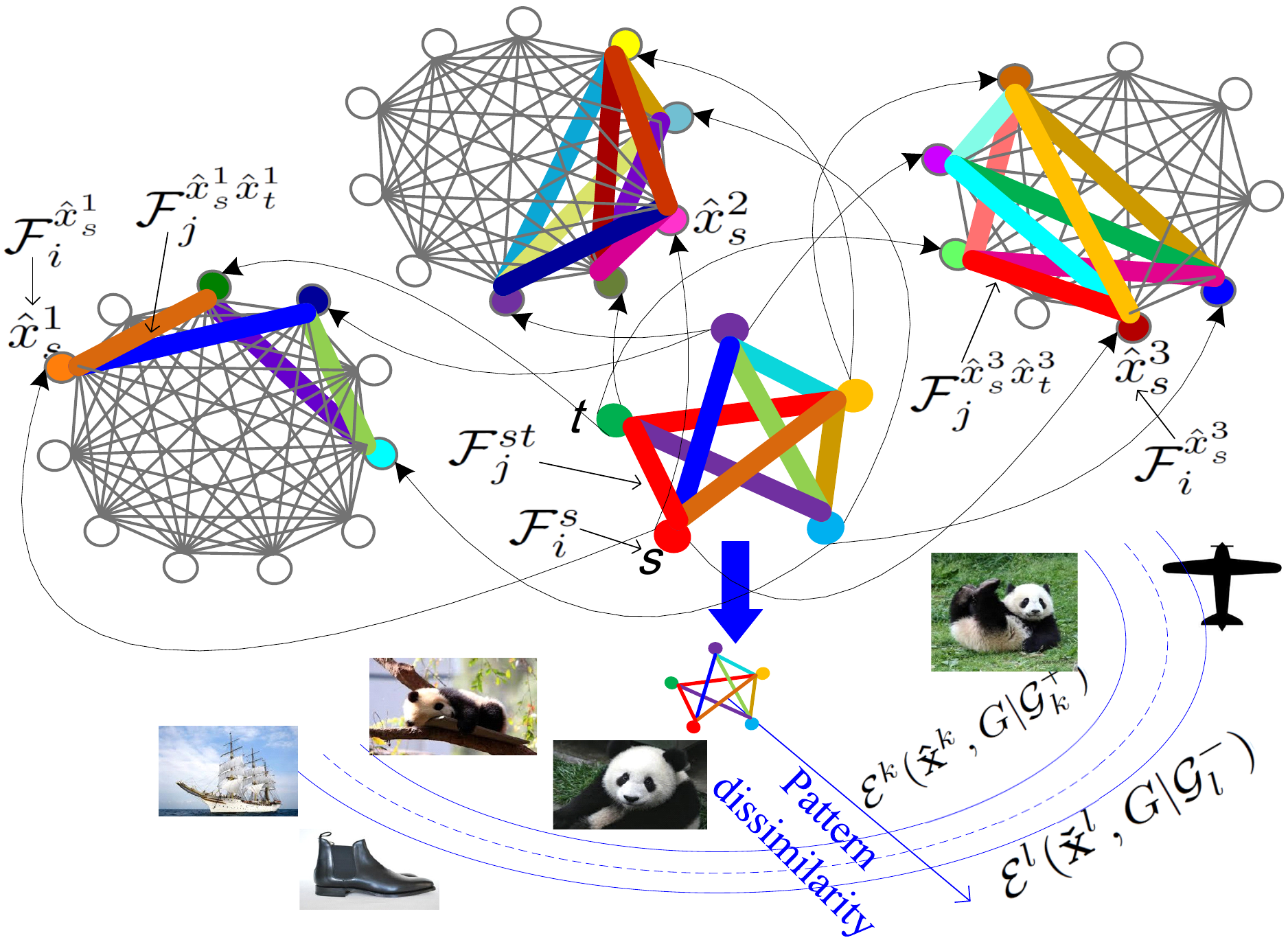}
\caption{Notation of the mVAP defined in Definition~\ref{def:VAP}}
\label{fig:VAP}
\end{figure}

\begin{definition}{{\bf(ARG)}}
An ARG ${\mathcal G}$ is a three-element tuple ${\mathcal G}=({\mathcal V},{\bf F}_{\mathcal V},{\bf F}_{\mathcal E})$, where ${\mathcal V}$ is the node set. Undirected edges connect each pair of nodes to form a complete graph. ${\mathcal G}$ contains $N_{P}$ types of local attributes for each node and $N_{Q}$ types of pairwise attributes for each edge. ${\bf F}_{\mathcal V}=\{{\mathcal F}_{i}^{x}|x\in{\mathcal V}, i=1,2,...,N_{P}\}$ and ${\bf F}_{\mathcal E}=\{{\mathcal F}_{j}^{x_{1}x_{2}}|x_{1},x_{2}\in{\mathcal V}, x_{1}\not=x_{2}, j=1,2,...,N_{Q}\}$ denote the local and pairwise attribute sets, respectively. Each attribute corresponds to a feature vector.
\label{def:ARG}
\end{definition}
Similar to the above notation, the maximal frequent subgraph pattern among positive ARGs ${\bf G}^{+}\!=\!\{{\mathcal G}^{+}_{k}|k\!=\!1,\!2,\!...,N^{+}\}$ can be represented as a five-element tuple $G\!=\!(V,E,{\bf F}_{V},{\bf F}_{E},{\bf W})$, and allows this to be an incomplete graph. Edges $(s,t)\!\in\!E$ and $(t,s)\!\in\!E$ are regarded as two different directed edges in $G$. ${\bf W}\!=\!\{w^{P}_{i}|i=1,2,...,N_{P}\}\cup\{w^{Q}_{j}|j=1,2,...,N_{Q}\}\cup\{P_{none},Q_{none}\}$ is the parameter set, where $w^{P}_{i}$ and $w^{Q}_{j}$ denote positive weights for local attributes ${\{{\mathcal F}_{i}^{s}|s\!\in\!V\}}$ and pairwise attributes ${\{{\mathcal F}_{i}^{st}|(s,t)\!\in\!E\}}$.

Then, we use a set of labels $\{\hat{x}_{s}^{k}|s\!\in\!V\}$ to represent the node correspondences between pattern $G$ and each positive ARG ${\mathcal G}_{k}^{+}$. We map each node $s$ in $G$ to an ARG node $\hat{x}_{s}^{k}\!\in\!{\mathcal V}_{k}^{+}$. Note that \textbf{occlusion} in ARGs should be also considered. Without loss of generality, $s$ is mapped to a dummy node \textit{none} ($\hat{x}_{s}^{k}\!\in\!{\mathcal V}_{k}^{+}\cup\{none\}$), when its corresponding node in ${\mathcal G}_{k}^{+}$ is \textbf{occluded}. Parameters $P_{none},Q_{none}\!\in\!{\bf W}$ are constant penalties for mapping $s$ to $none$. Typically, we use square differences to define the attribute dissimilarity (or fuzziness) of matching $s$ to $\hat{x}_{s}^{k}$ in ARG ${\mathcal G}_{k}^{+}$.
\begin{small}
\begin{equation}
\begin{split}
{\mathcal E}_{s}^{k}(\hat{x}_{s}^{k},G|{\mathcal G}_{k}^{+})\!=\!P_{s}(\hat{x}_{s}^{k},G|{\mathcal G}_{k}^{+})\!+\!{\sum}_{(s,t)\in{E_{s}}}\!\!Q_{st}(\hat{x}_{s}^{k},\hat{x}_{t}^{k},G|{\mathcal G}_{k}^{+})\!\!\!\!\!\!\!\!\!\\
\!\!\!\!\!P_{s}(\hat{x}_{s}^{k},G|{\mathcal G}_{k}^{+})=\left\{\!\!
\begin{array}{ll}\sum^{N_{P}}_{i=1}\!w^{P}_{i}\Vert{\mathcal F}_{i}^{s}\!-\!{\mathcal F}_{i}^{\hat{x}_{s}^{k}}\Vert^2\!\!,&\!\!\hat{x}_{s}^{k}\!\!\in\!\!{\mathcal V}_{k}^{+}\\P_{none},&\!\!\!\!\!\!\!\!\hat{x}_{s}^{k}\!=\!none\end{array}\right.\!\!\!\!\!\\
\!\!\!\!\!Q_{st}(\hat{x}_{s}^{k},\hat{x}_{t}^{k},G|{\mathcal G}_{k}^{+})=\left\{\!\!\begin{array}{ll}{\sum^{N_{Q}}_{j=1}\!w^{Q}_{j}\Vert{\mathcal F}_{j}^{st}\!\!-\!{\mathcal F}_{j}^{\hat{x}_{s}^{k}\hat{x}_{t}^{k}}\!\Vert^2}/{\vert{E_{s}}\vert},
\!\!\!\!\!\!\!\!\!\!\!\!\!\!\!\!\!\!\!
\!\!\!\!\!\!\!\!\!\!\!\!\!\!\!\!\!\!\!
\!\!\!\!\!\!\!\!\!\!\!\!\!\!\!\!\!\!\!&\\
&\hat{x}_{s}^{k}\!\not=\!\hat{x}_{t}^{k}\!\!\in\!\!{\mathcal V}_{k}^{+}\\+\infty,&\!\!\!\hat{x}_{s}^{k}=\hat{x}_{t}^{k}\in{\mathcal V}_{k}^{+}\\{Q_{none}}/{\vert{E_{s}}\vert},&\!\!\!\!\!\hat{x}_{s}^{k}\,\textrm{or}\,\hat{x}_{t}^{k}\!=\!none\end{array}\right.
\end{split}
\label{eqn:energy}
\end{equation}
\end{small}
where functions $P_{s}$ and $Q_{st}$ measure the difference in local and pairwise attributes. Infinite penalties are used to avoid many-to-one node assignments.
\begin{definition}{{\bf(mVAP)}} Given a set of positive ARGs ${\bf G}^{+}\!=\!\{{\mathcal G}^{+}_{k}|k\!=\!1,\!2,\!...,N^{+}\}$, a set of negative ARGs ${\bf G}^{-}\!=\!\{{\mathcal G}^{-}_{k}|k\!=\!1,\!2,\!...,N^{-}\}$, the minimum degree $d$, and a threshold $\tau$, $G\!=\!(V,\!E,\!{\bf F}_{V}\!,\!{\bf F}_{E}\!,\!{\bf W})$ is a mVAP, if and only if:\\
\textbf{(a)} ${\min}_{V,\{\hat{x}_{s}^{k}\},{\bf F}_{V},{\bf F}_{E}}\sum_{s\in V}[{\mathcal E}_{s}({\bf\hat{x}}_{s},G|{\bf G}^{+})-\tau]$,\\
where ${\mathcal E}_{s}({\bf\hat{x}}_{s},G|{\bf G}^{+})\!=\!\textrm{mean}_{k=1}^{N^{+}}{\mathcal E}_{s}^{k}(\hat{x}_{s}^{k},G|{\mathcal G}_{k}^{+})$;\\
\textbf{(b)} $\forall s\in V, {\max}_{E_{s}:\vert{E_{s}}\vert\geq\min\{d,\vert{V}\vert-1\},{\mathcal E}_{s}({\bf\hat{x}}_{s},G|{\bf G}^{+})<\tau}\vert{E_{s}}\vert$;\\
\textbf{(c)} $\underset{{\bf W}}{\min}\Vert{\bf w}\Vert^2+\frac{C}{N^{+}}\sum_{k=1}^{N^{+}}{\xi}_{k}^{+}+\frac{C}{N^{-}}\sum_{l=1}^{N^{-}}{\xi}_{l}^{-}$,\\
{\verb|  |}${\forall k=1,2,...,N^{+}, -[{\mathcal E}^{k}({\bf\hat{x}}^{k},G|{\mathcal G}_{k}^{+})+b]\!\geq\!1\!-\!{\xi}_{k}^{+}}$,\\
{\verb|  |}${\forall l=1,2,...,N^{-}, {\mathcal E}^{l}({\bf\check{x}}^{l},G|{\mathcal G}_{l}^{-})+b\!\geq\!1\!-\!{\xi}_{l}^{-}}$,\\
where ${\bf\check{x}}^{l}\!\leftarrow\!{\arg\!\min}_{{\bf\check{x}}^{l}}{\mathcal E}^{l}({\bf\check{x}}^{l},G|{\mathcal G}_{l}^{-})$,\\
{\verb|  |}${\mathcal E}^{k}({\bf\hat{x}}^{k},G|{\mathcal G}_{k}^{+})\!=\!{\sum}_{s\in V}{\mathcal E}_{s}^{k}(\hat{x}_{s}^{k},G|{\mathcal G}_{k}^{+})$.
\label{def:VAP}
\end{definition}

\textbf{Item~(a)} presents a basic principle for a subgraph pattern among positive ARGs. It estimates the best node correspondences $\{{\bf\hat{x}}_{s}\}$ and attributes $({\bf F}_{V},{\bf F}_{E})$ for $G$ that minimize the overall pattern fuzziness. In other words, this item requires the pattern and its corresponding subgraphs in different ARGs to have similar attributes, where the average attribute dissimilarity between each node $s$ and its corresponding nodes is given by ${\mathcal E}_{s}({\bf\hat{x}}_{s},G|{\bf G}^{+})$. The attributes and node correspondences are estimated via global optimization, considering large intra-category attribute variations. In addition, to ensure low fuzziness in pattern $G$, we use a threshold $\tau$ to limit ${\mathcal E}_{s}({\bf\hat{x}}_{s},G|{\bf G}^{+})$. Note that this also maximizes the pattern size $\vert V\vert$, as long as the fuzziness ${\mathcal E}_{s}({\bf\hat{x}}_{s},G|{\bf G}^{+})$ for each node $s$ is less than $\tau$. $E_{s}\!=\!\{(s,t)|t\!\in\!V\}\!\subset\!E$ denotes the set of outgoing edges of $s$. ${\bf\hat{x}}_{s}\!=\!\{\hat{x}_{s}^{k}|k\!=\!1,2,...,N^{+}\}$ denotes matching assignments of $s$.

\textbf{Item~(b)} mines the hidden node dependency in $G$. Given a certain fuzziness limit $\tau$ for $G$, this item encourages each pattern node $s$ to have as many outgoing edges as possible. In addition, a minimum edge number $d$ is used to ensure a certain pattern density.

\textbf{Item~(c)} discovers latently effective metrics of pattern similarity for $G$. Different attributes in $G$ correspond to different metrics of distance measurements, and the effective ones should have a good capacity for classifying positive and negative\footnote{Negative ARGs do not contain pattern $G$ and usually represent background.} subgraphs. ${\mathcal E}^{k}({\bf\hat{x}}^{k},G|{\mathcal G}_{k}^{+})$ measures the average attribute dissimilarity between $G$ and the corresponding subgraph in ${\mathcal G}_{k}^{+}$, where ${\bf\hat{x}}^{k}\!=\!\{\hat{x}_{s}^{k}|s\!\in\!V\}$.

\section{Graph mining}
\label{sec:mining}

\begin{figure}
\centering
\includegraphics[width=\linewidth]{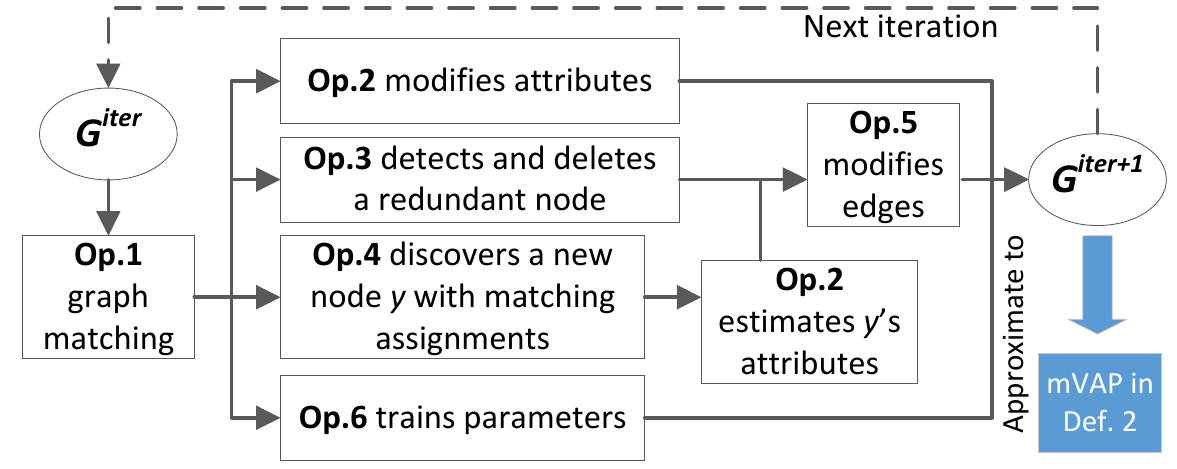}
\caption{Flowchart of an approximate solution to the NP-hard mining problem}
\label{fig:flowchart}
\end{figure}

Given a set of positive ARGs ${\bf G}^{+}$, a set of negative ARGs ${\bf G}^{-}$, and an initial graph template $G^{0}$ that roughly corresponds to a fragment of the target subgraph pattern, the goal of graph mining is to iteratively estimate all the parameters contained in the pattern and modify $G^{0}$ to the maximal-size VAP among these ARGs, $G^{0}\!\rightarrow\!G^{1}\!\rightarrow\!...\!\rightarrow\!G^{n}\!=\!\textrm{mVAP}$. Therefore, as shown in Fig.~\ref{fig:flowchart}, we define six operations, which form an EM framework, to modify the current pattern. We can demonstrate\footnote{Please see the supplementary materials for proofs.} that these operations present an approximate but efficient solution to graph mining.

\noindent\textbf{Initialization:} The initial graph template $G^{0}$ can be manually labeled as a complete graph. Even bad labeling is acceptable, \emph{e.g.} using an object fragment mixed with background parts to construct $G^{0}$. We then initialize the matching parameters in ${\bf W}^{0}$ as $P_{none}^{0}\!=\!Q_{none}^{0}\!=\!+\infty$ and $w^{P,0}_{i\!=\!1,2,...,N_{P}}\!=\!w^{Q,0}_{j\!=\!1,2,...,N_{Q}}\!=\!1/(N_{P}+N_{Q})$.

\noindent\textbf{Op. 1, graph matching:} Node correspondences are estimated using Definition~\ref{def:VAP}(a), in which the energy is written as
\begin{small}
\begin{equation}
\textrm{Energy}^{\textrm{(a)}}({\bf\hat{x}},G)\!=\!\sum_{s\in V}[{\mathcal E}_{s}({\bf\hat{x}}_{s},G|{\bf G}^{+})-\tau], \textrm{where}\;{\bf\hat{x}}\!=\!\bigcup_{s\in V}{\bf\hat{x}}_{s}\!\!\!
\end{equation}
\end{small}
We update node correspondences between $G^{iter}$ and positive ARGs in a new iteration.
\begin{small}
\begin{equation}
\begin{split}
\!&\!\forall k,\;\{\hat{x}_{s}^{k}\}^{iter+1}\!\!\!\!\leftarrow\!{\arg\!\min}_{\{\hat{x}_{s}^{k}\}}\textrm{Energy}^{\textrm{(a)}}({\bf\hat{x}},G^{iter})\\
\!&\!=\!{\arg\!\min}_{\{\hat{x}_{s}^{k}\}}{\sum}_{s\in V^{iter}}{\mathcal E}_{s}^{k}(\hat{x}_{s}^{k},G^{iter}|{\mathcal G}_{k}^{+})\\
\!&\!=\underset{\{\hat{x}_{s}^{k}\}}{\arg\!\min}\!\!\!\!\sum\limits_{s\in V^{iter}}\!\!\!\!P_{s}(\hat{x}_{s}^{k},G^{iter}|{\mathcal G}_{k}^{+})\!+\!\!\!\!\!\!\!\!\!\!\sum\limits_{(s,t)\in{E^{iter}}}\!\!\!\!\!\!\!\!Q_{st}(\hat{x}_{s}^{k},\hat{x}_{t}^{k},G^{iter}|{\mathcal G}_{k}^{+})\!\!\!
\end{split}
\label{eqn:matching}
\end{equation}
\end{small}
This quadric assignment problem (QAP) is a typical case of graph matching, and can be solved by global optimization techniques, such as \cite{TRWS}.

\noindent\textbf{Op. 2, attribute estimation:} Based on Definition~\ref{def:VAP}(a), we update the attributes of $G^{iter}$ in a new iteration.
\begin{small}
\begin{equation}
({\bf F}_{V}^{iter+1}\!\!,{\bf F}_{E}^{iter+1})\!\leftarrow\!{\arg\!\min}_{{\bf F}_{V},{\bf F}_{E}}\,\textrm{Energy}^{\textrm{(a)}}({\bf\hat{x}}^{iter+1},G^{iter})\!\!\!
\end{equation}
\end{small}
For clarity, we simply use $\hat{x}_{s}^{k}$ to denote the node correspondence in the $(iter+1)$ iteration. The above equation can be solved as
\begin{small}
\begin{equation}
{\mathcal F}_{i}^{s,iter+1}=\!\!\!\!\!\underset{k:\delta(\hat{x}_{s}^{k})=1}{\textrm{mean}}\!\!\!{{\mathcal F}_{i}^{\hat{x}_{s}^{k}}},\quad{\mathcal F}_{j}^{st,iter+1}=\!\!\!\!\!\!\!\underset{k:\delta(\hat{x}_{s}^{k})\delta(\hat{x}_{t}^{k})=1}{\textrm{mean}}\!\!\!\!\!{{\mathcal F}_{j}^{\hat{x}_{s}^{k}\hat{x}_{t}^{k}}}
\label{eqn:attribute}
\end{equation}
\end{small}
where $\delta(\cdot)$ indicates whether a node is matched to $none$. If $\hat{x}^{k}_{s}\!=\!none$, then $\delta(\hat{x}^{k}_{s})\!=\!0$; otherwise, $1$.

\noindent\textbf{Op. 3, delete a redundant node:} According to Definition~\ref{def:VAP}(a), this operation selects and deletes the worst node $\hat{s}$ from $G^{iter}$ to reduce the overall energy.
\begin{small}
\begin{equation}
\begin{split}
&\Delta_{s}^{\textrm{del}}\textrm{Energy}^{\textrm{(a)}}({\bf\hat{x}}^{iter+1},G^{iter})\!=\!\tau\!-\!{\mathcal E}_{s}({\bf\hat{x}}_{s}^{iter+1}\!,\!G^{iter}|{\bf G}^{+})\\
&\hat{s}\!\leftarrow\!{\arg\!\min}_{s\in V^{iter}}\Delta_{s}^{\textrm{del}}\textrm{Energy}^{\textrm{(a)}}({\bf\hat{x}}^{iter+1},G^{iter})\\
&\textrm{if}\;\Delta_{\hat{s}}^{\textrm{del}}\textrm{Energy}^{\textrm{(a)}}({\bf\hat{x}}^{iter+1}\!\!,\!G^{iter})\!<\!0\;\textrm{then}\;V^{iter+1}\!\!\!\leftarrow\!V^{iter}\!\!\setminus\!\{\hat{s}\}\!\!\! \end{split}
\end{equation}
\end{small}
The energy change from removing node $s$ is denoted by ${\Delta_{s}^{\textrm{del}}\textrm{Energy}^{\textrm{(a)}}({\bf\hat{x}}^{iter+1},G^{iter})}$, which is actually affected by the connection of $s$, $E_{s}$. Thus, we need to take linkage estimation in Definition~\ref{def:VAP}(b) into account. Therefore, in this operation, we tentatively assign each node $s$ with the linkages $E_{s}$ that minimize ${{\mathcal E}_{s}({\bf\hat{x}}_{s}^{iter+1}\!,\!G^{iter}|{\bf G}^{+})}$ to protect the good nodes from being deleted, \emph{i.e.} ${\arg\!\min}_{E_{s}:\vert{E_{s}}\vert\!\geq\!d_1}{\mathcal E}_{s}({\bf\hat{x}}_{s}^{iter+1}\!,\!G^{iter}|{\bf G}^{+})\!=\!\{(s,t)\!\in\!E^{iter}|1\!\leq\!{{\textrm{rank}}}_{t\in V^{iter}}{\sum}_{k=1}^{N^{+}}Q_{st}(\hat{x}_{s}^{k},\!\hat{x}_{t}^{k},\!G^{iter}|{\mathcal G}_{k}^{+})\!\leq\!d_1\}$, where $d_1\!=\!\min\{d,\vert{V^{iter}}\vert-1\}$. Thus, based on $\{E_{s}\}$ and ${\bf\hat{x}}$, we can directly compute $\hat{s}$ in the above equation.

\noindent\textbf{Op. 4, node discovery:} We formulate the energy changes considering both Definition~\ref{def:VAP}(a) and Definition~\ref{def:VAP}(b):
\begin{small}
\begin{equation}
\begin{split}
\!\!&\Delta_{y}^{\textrm{add}}\textrm{Energy}^{\textrm{(a)}}({\bf\hat{x}}^{iter+1}\!\!,\!G^{iter})\!=\!\!\!\!\!\!\!\!\!\!\!\!\!\!\!\!\min_{\{\hat{x}_{y}^{k}\},\{{\mathcal F}_{i}^{y}\},\{{\mathcal F}_{j}^{yt}\},E_{y}}\!\!\!\!\!\!\!\!\!\!{\mathcal E}_{y}({\bf\hat{x}}_{y},G^{new}|{\bf G}^{+})\!-\!\tau\!\!\!\!\!\\
\!\!&\textrm{if}\;\Delta_{y}^{\textrm{add}}\textrm{Energy}^{\textrm{(a)}}({\bf\hat{x}}^{iter+1}\!\!,\!G^{new})\!<\!0\;\textrm{then}\;V^{iter+1}\!\!\!\leftarrow\!V^{iter}\!\bigcup\{y\}\!\!\!\!\!
\end{split}
\end{equation}
\end{small}
where $y$ denotes a missing node of $G^{iter}$, and $G^{new}$ corresponds to a dummy enlarged pattern including $y$. In this operation, we need to simultaneously discover matching assignments $\{\hat{x}_{y}^{k}\}$ and attributes $(\{{\mathcal F}_{i}^{y}\},\{{\mathcal F}_{j}^{yt}\})$ of the new node $y$ that represent the most reliable hidden pattern in ARGs. We have the following approximate solution to this problem. First, we use the following equation to estimate rough values of $\{\hat{x}_{y}^{k}\}$, which can be regarded as a QAP of a Markov random field (MRF) \emph{w.r.t} $\{\hat{x}_{y}^{k}\}$ and directly solved.
\begin{small}
\begin{equation}
\begin{split}
\!&\qquad{\arg\!\min}_{\{\hat{x}_{y}^{k}\}}{\sum}_{1\!\leq\!k,l\!\leq\!N^{+}}\!\!\!\widetilde{M}_{kl}(\hat{x}_{y}^{k},\hat{x}_{y}^{l})\\
\!&\textrm{where}\,\widetilde{M}_{kl}(\hat{x}_{y}^{k},\hat{x}_{y}^{l})\!\!=\!\!\!\!\!\underset{\vert{E_{y}}\vert\!=\!d_2}{\min}\!\!\!\!\!\sum\limits_{(y,t)\!\in\!E_{y}}\!\!\!\!\!m^{kl}_{t}\!+\!\!\sum^{N_{P}}_{i=1}\frac{w^{P}_{i}\Vert{\mathcal F}_{i}^{\hat{x}_{y}^{k}}-{\mathcal F}_{i}^{\hat{x}_{y}^{l}}\Vert^2}{2(N^{+})^2}\!\!\!\\
\!&m^{kl}_{t}\!=\!\frac{\delta(\hat{x}^{k}_{t})\delta(\hat{x}^{l}_{t})\sum^{N_{Q}}_{i=1}w^{Q}_{i}\Vert{\mathcal F}_{i}^{\hat{x}_{y}^{k}\hat{x}^{k}_{t}}-{\mathcal F}_{i}^{\hat{x}_{y}^{l}\hat{x}^{l}_{t}}\Vert^2}{2\vert{E_{y}}\vert(N^{+})\sum_{j}\delta(\hat{x}^{j}_{t})}\\
\!&\qquad+\frac{(1-\delta(\hat{x}^{k}_{t})\delta(\hat{x}^{l}_{t}))Q_{none}}{\vert{E_{y}}\vert(N^{+})(N^{+}+\sum_{j}\delta(\hat{x}^{j}_{t}))}
\end{split}
\label{eqn:new}
\end{equation}
\end{small}
Second, we use $\{\hat{x}_{y}^{k}\}$ to estimate $E_{y}$ as $\{(y,t)|t\in V^{iter},1\!\leq\!{\textrm{rank}}_{t\in V^{iter}}{\sum}_{k=1}^{N^{+}}Q_{yt}(\hat{x}_{y}^{k},\hat{x}_{t}^{k},G^{iter}|{\mathcal G}_{k}^{+})\!\leq\!d_2\}$, where $d_2\!=\!\min\{d,\vert{V}\vert\}$. We avoid the uncertainty in node linkages of $y$, $E_{y}$, by tentatively applying the most conservative setting (accurate linkages will be estimated in \textbf{Op. 5}). Third, given $E_{y}$, we can further refine matching assignments $\{\hat{x}_{y}^{k}\}$ as follows.
\begin{small}
\begin{equation}
\begin{split}
\!&\qquad{\arg\!\min}_{\{\hat{x}_{y}^{k}\}}{\sum}_{1\leq k,l\leq N^{+}}M_{kl}(\hat{x}_{y}^{k},\hat{x}_{y}^{l})\\
\!&M_{kl}(\hat{x}_{y}^{k},\hat{x}_{y}^{l})\!=\!\frac{\sum^{N_{P}}_{i=1}w^{P}_{i}\Vert{\mathcal F}_{i}^{\hat{x}^{k}_{y}}-{\mathcal F}_{i}^{\hat{x}^{l}_{y}}\Vert^2}{2(N^{+})^2}\!+\!{\sum}_{t:(y,t)\in E_{y}}\!m^{kl}_{t}\!\!\!\!\!\!\!\!\!
\end{split}
\label{eqn:new_refine}
\end{equation}
\end{small}
Finally, given the refined $\{\hat{x}_{y}^{k}\}$, $y$'s attributes $(\{{\mathcal F}_{i}^{y}\},\{{\mathcal F}_{j}^{yt}\})$ can be further estimated using \textbf{Op. 2}.

\noindent\textbf{Op. 5, fill edges:} Based on Definition~\ref{def:VAP}(b), we design the following procedure to update the edge set of each node $s$.

\begin{algorithmic}
\STATE {\bfseries Initialization} $E_{s}^{iter+1}=\emptyset$.
\FOR{$i=1$ {\bfseries to} $\vert{V^{iter}}\vert-1$}
\STATE $\textrm{{\bf If}}\;{\mathcal E}_{s}({\bf\hat{x}}_{s},G|{\bf G}^{+})|_{E_{s}^{iter+1}\bigcup\{(s,t_{i})\}}<\tau$ $\textrm{{\bf then}}$ $E_{s}^{iter+1}\!\!\!\!\leftarrow\!E_{s}^{iter+1}\!\bigcup\{(s,t_{i})\}$, $\textrm{{\bf else} break;}$ $\textrm{where}\;t_{i}\!=\!{\arg}_{t\in V}\big($ $\textrm{rank}{\sum}_{k=1}^{N^{+}}Q_{st}(\hat{x}_{s}^{k}, \hat{x}_{t}^{k},G^{iter}|{\mathcal G}_{k}^{+})\!=\!i\big)$.
\ENDFOR
\end{algorithmic}

\begin{figure}
\centering
\includegraphics[width=\linewidth]{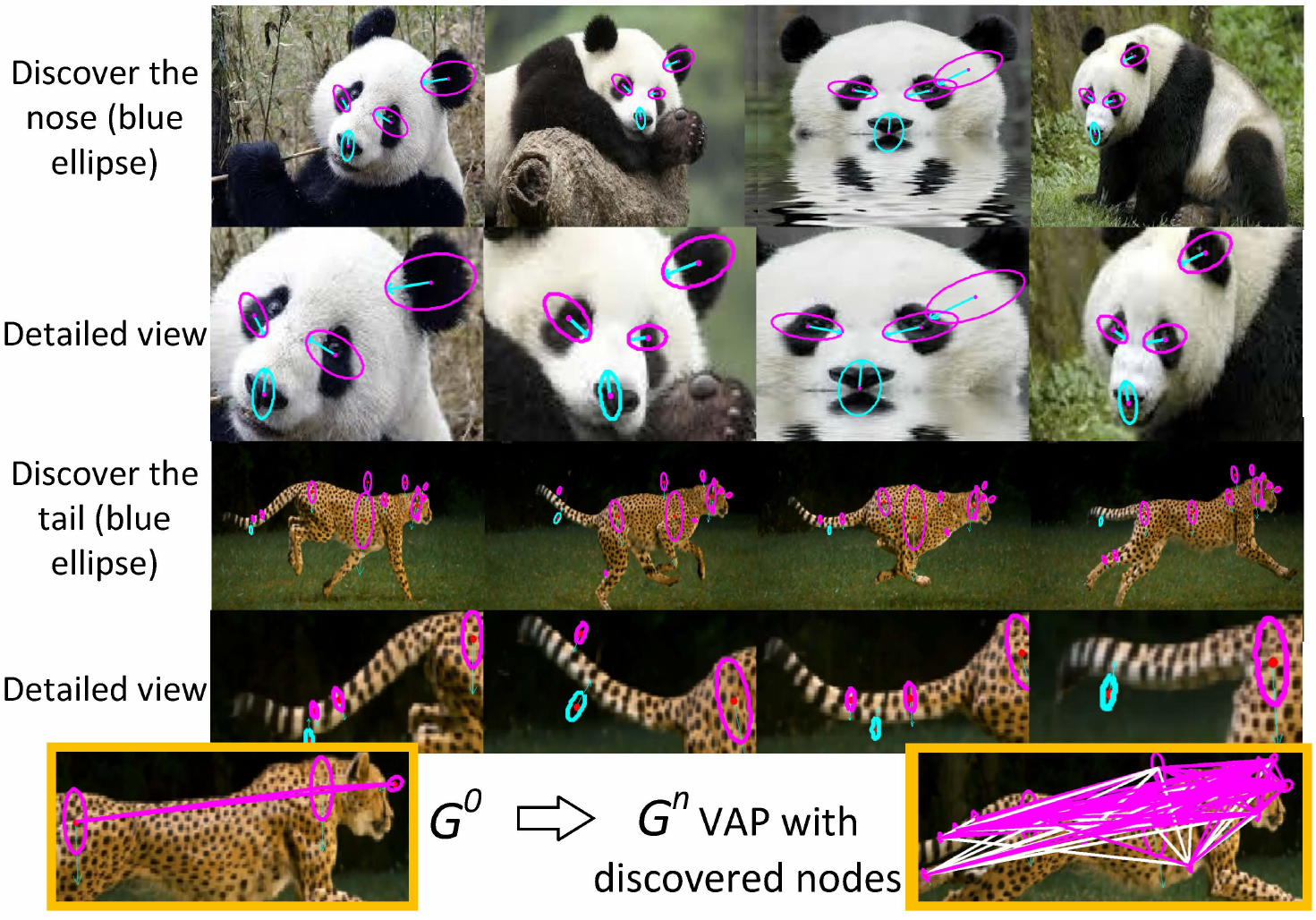}
\caption{Node discovery. The panda's nose and the cheetah's tail (cyan) are identified as the most probably missing nodes and added to $G^{iter}$ (magenta).}
\label{fig:new}
\end{figure}

\noindent\textbf{Op. 6, train matching parameters:} Parameter training involved in Definition~\ref{def:VAP}(b) can be solved using a linear SVM.
\begin{small}
\begin{equation}
\begin{split}
\!\!\!\min_{{\bf w},{\bf\xi},b}\Vert{\bf w}\Vert^2+\frac{C}{N^{+}}{\sum}^{N^{+}}_{k=1}\!\xi^{+}_{k}+\frac{C}{N^{-}}{\sum}^{N^{-}}_{l=1}\!\xi^{-}_{l},\!\!\!\!\!\\
\!\!\!\!\!\textrm{s.t.}\; \forall k\!=\!1,2,...,N^{+}\!,-({\bf w}\!\cdot\!{\bf a}^{+}_{k}\!\!-\!b)\!\!\geq\!\! 1\!-\!\xi^{+}_{k},\; \xi^{+}_{k}\!\geq\! 0;\\
\forall l\!=\!1,2,...,N^{-}\!,({\bf w}\!\cdot\!{\bf a}^{-}_{l}\!\!-\!b)\!\!\geq\!\! 1\!-\!\xi^{-}_{l},\; \xi^{-}_{l}\!\geq\!0
\end{split}
\label{eqn:svm}
\end{equation}
\end{small}
where ${\bf w}\!=\![w^{P}_{1},...,w^{P}_{N_{P}},w^{Q}_{1},...,w^{Q}_{N_{Q}}]^{T}, {\bf a}^{+}_{k}\!=\![a^{k,P}_{1},...$, $a^{k,P}_{N_{P}},a^{k,Q}_{1},...,a^{k,Q}_{N_{Q}}]^{T}$, $a^{k,P}_{i}\!=\!\textrm{mean}_{s\!\in\!V:\delta(\hat{x}^{k}_{s})\!=\!1}$ $\Vert{\mathcal F}_{i}^{s}\!-\!{\mathcal F}_{i}^{\hat{x}^{k}_{s}}\Vert^2$, $a^{k,Q}_{j}\!=\!\textrm{mean}_{s\!\in\!V:\delta(\hat{x}^{k}_{s})\!=\!1}\textrm{mean}_{(s,t)\!\in\!E_{s}:\delta(\hat{x}^{k}_{t})\!=\!1}$ ${\Vert{\mathcal F}_{j}^{st}\!-\!{\mathcal F}_{j}^{\hat{x}^{k}_{s}\hat{x}^{k}_{t}}\Vert^2}$. Given ${\bf\check{x}}^{l}$\footnote{We tentatively set $P_{none}$ and $Q_{none}$ to be infinite to avoid matching to $none$ in the calculation of ${\bf\check{x}}^{l}$.}, we can compute ${\bf a}^{-}_{l}$ in the same way as ${\bf a}^{+}_{k}$.

Computing ${\bf w}$ as above, we remove all the negative elements in ($w_{i}\!\leftarrow\!\max\{w_{i},0\}$) and normalize ${\bf w}$ (${\bf w}\!\leftarrow\!{\bf w}/\Vert{\bf w}\Vert_{1}$). We gradually update the attribute weights in each iteration, \emph{i.e.} ${\bf w}^{iter+1}\!\!\leftarrow\!\lambda{\bf w}\!+\!(1-\lambda){\bf w}^{iter}$, where $\lambda\!=\!0.5$. Finally, given ${\bf w}^{iter+1}$, we estimate $P_{none}^{iter+1}$ and $Q_{none}^{iter+1}$ as
\begin{small}
\begin{equation}
\begin{split}
\!\!&P_{none}^{iter+1}\!\leftarrow\!\bar{P}^{+}\!\!+\!\alpha(\bar{P}^{-}\!\!-\!\bar{P}^{+}\!),\; Q_{none}^{iter+1}\!\leftarrow\!\bar{Q}^{+}\!\!+\!\alpha(\bar{Q}^{-}\!\!-\!\bar{Q}^{+}\!);\!\!\!\!\!\!\!\!\!\!\!\!\!\\
\!\!&\bar{P}^{+}\!\!=\!{\textrm{mean}}_{1\!\leq\!k\!\leq\!N^{+}\!\!,s\in\!V^{iter}}P_{s}(\hat{x}_{s}^{k},G^{iter}|{\mathcal G}_{k}^{+}),\\
\!\!&\bar{Q}^{+}\!\!=\!{\textrm{mean}}_{1\!\leq\!k\!\leq\!N^{+}\!\!,s\!\in\!V^{iter}}{\sum}_{(s,t)\!\in\!E_{s}}Q_{st}(\hat{x}_{s}^{k},\hat{x}_{t}^{k},G^{iter}|{\mathcal G}_{k}^{+})\!\!\!\!\!\!\!\!\!\!\!\!\!\!
\end{split}
\end{equation}
\end{small}
where $w^{Q}_{j}$, $w^{P}_{i}$ are elements in ${\bf w}$; $\bar{P}^{+}$, $\bar{Q}^{+}$ denote the average unary and pairwise matching penalties in positive ARGs, respectively; $\bar{P}^{-}$, $\bar{Q}^{-}$ for negative ARGs are defined in the same way as $\bar{P}^{+}$, $\bar{Q}^{+}$; and $\alpha>0$ (we set $\alpha\!=\!1.0$).

\section{Experiments}
\label{sec:exper}

The proposed visual graph mining provides a general solution to the discovery of common patterns from cluttered visual data, where the target objects are randomly placed. Therefore, we design five experiments and test the generality of our method by applying it to four types of visual data, including unlabeled RGB-D and RGB indoor scenes, web images, and videos. Four kinds of ARGs are designed to represent these data.

In the first two experiments, we apply our method to cluttered indoor RGB-D and RGB images, respectively. We use two types of ARGs, which use edge segments as graph nodes, to represent the indoor objects with clear edges. Then, in Experiments 3 and 4, we apply our method to more general images, \emph{i.e.} ubiquitous web images that can be collected from the internet and those in the challenging Pascal VOC dataset, respectively. Consequently, we design another two types of ARGs for image representation. One type of ARGs use SIFT feature points as graph nodes, and the other type of ARGs choose the automatically extracted middle-level patches as nodes. Finally, we further test the performance of mining deformable models from videos in Experiment 5.

We design a total of seven competing methods to enable a comprehensive comparison. They include typical image/graph matching approaches, unsupervised methods for learning graph matching, and the only pioneering method in the scope of visual graph mining~\cite{OurCVPR14Graph}, which consider both the expressive power of graph mining and the challenges of visual mining.

\begin{figure}
\centering
\includegraphics[width=\linewidth]{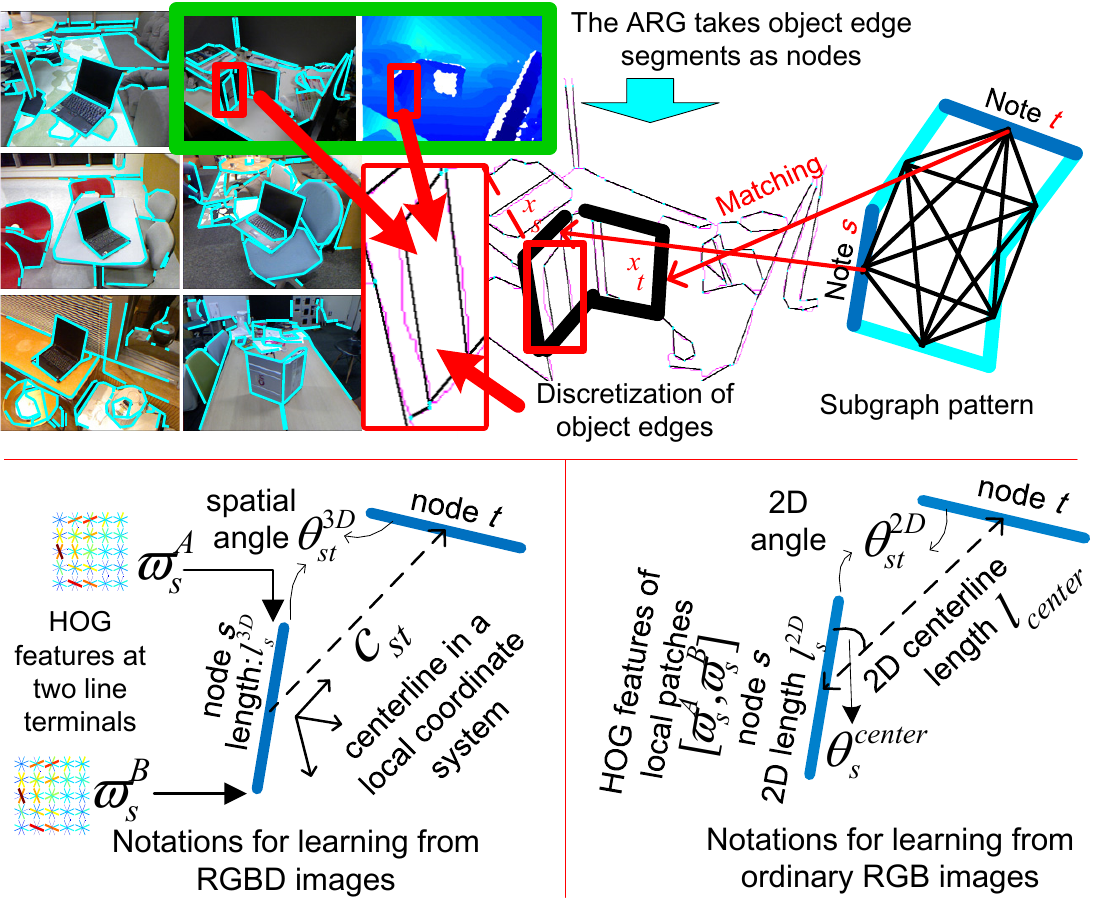}
\caption{Notation for edge-based ARGs that are used in Experiments 1 and 2}
\label{fig:ARG_edge}
\end{figure}

\subsection{Experiment 1: mining edge-based models from cluttered indoor RGB-D images}

\textbf{Dataset:}{\verb| |} The Kinect RGB-D image dataset proposed in \cite{OurCVPR} is the benchmark RGB-D image dataset for testing graph matching and mining\footnote{This is one of the largest RGB-D object datasets. This dataset was designed for the evaluation of graph matching and graph mining, as it contains multiple kinds of intra-category variation, such as variations in texture, rotation, scale, and structure deformation.}, and has been used in \cite{OurCVPR14Graph,OurICCV13}. In this study, we use six categories, \emph{i.e.} \textit{notebook PC}, \textit{drink box}, \textit{basket}, \textit{bucket}, \textit{sprayer}, and \textit{dustpan}, which contain sufficient objects for both training and testing. All the target objects are randomly placed in cluttered indoor scenes.

\textbf{Edge-based ARGs for RGB-D images:}{\verb| |} We use edge-based ARGs, which were widely applied in \cite{OurCVPR,OurICCV13,OurCVPR14Graph}, to represent RGB-D objects in indoor environment. The ARGs are designed as follows. First, object edges are extracted from the images using the technique of \cite{EdgeDetection}, which uses only the RGB channels in the RGB-D images. Then, continuous edges are divided into line segments, and we take these line segments as graph nodes of the ARGs, as illustrated in Fig.~\ref{fig:ARG_edge}. Please see \cite{OurCVPR} for technical details of edge segmentation. Each pair of graph nodes is connected to construct a complete graph.

A total of two local attributes ($N_{P}=2$) and three pairwise attributes ($N_{Q}=3$) are designed to achieve robustness to rotation and scale variations in graph matching. As the first unary attribute, HoG features~\cite{HOG} are extracted from two patches at the line terminals of each node $s$. The HoG feature contains $5\times 5$ cells, each covering half of its neighbors. In each cell, gradients are computed using four orientation bins from $0^{\circ}$ to $180^{\circ}$. As the locally collected patch does not suffer much from illumination changes, all the cells are normalized within a single block. The second unary attribute describes the spatial length of the line segment of each node $s$ ($l^{3D}_{s}$). The attribute is defined as the logarithm of the line length ${\mathcal F}_{2}^{s}=\log l^{3D}_{s}$. The first pairwise attribute, which is given by ${\mathcal F}_{1}^{st}=\theta^{3D}_{st}$, represents the spatial angle between each pair of line segments $s$ and $t$. For each edge $(s,t)$, we define the line connecting the centers of the line segments $s$ and $t$ as the \textit{centerline} of $(s,t)$. This centerline can be regarded as the relative spatial translation between two nodes $s$ and $t$. To construct rotation-robust ARGs, \cite{OurCVPR} defined a local 3D coordinate system for each centerline that is independent of the global object rotation to measure the translation. Let ${\bf c}_{st}$ denote the translation between nodes $s$ and $t$. The second and third pairwise attributes represent the length and local orientation of the translation, and are written as ${\mathcal F}_{2}^{st}=\Vert{\bf c}_{st}\Vert$ and ${\mathcal F}_{3}^{st}={{\bf c}_{st}}/\Vert{\bf c}_{st}\Vert$, respectively.

\textbf{Experimental details:}{\verb| |} We test the graph-mining performance of the proposed method under different parameter settings of $d$ and $\tau$. We follow the same experimental settings as in \cite{OurCVPR14Graph}, including the same labeling of the initial graph templates $G^{0}$ and the same leave-one-out cross validation process.

Given each parameter setting, we perform a series of cross validations. For each RGB-D image in a category, we take the target object within it as the initial graph template and start an individual mining process. We randomly select $2/3$ and $1/3$ of the remaining images in this category to construct the positive ARGs for training and testing, respectively. We use images in the other categories to construct negative ARGs. We randomly select the same number of negative ARGs as positive ARGs in terms of both training and testing. In this way, given each specific parameter setting, we obtain a set of mVAPs for a category. In Section~\ref{sec:competing}, we propose a number of metrics for evaluation, and the overall performance of graph mining with certain parameters is evaluated by computing the average performance among all the mined mVAPs in all the categories.

\subsection{Experiment 2: mining edge-based models from cluttered indoor RGB images}

The design of Experiment 2 is similar to that of Experiment 1. We apply the same dataset of Kinect RGB-D images, but use only the RGB channels in the RGB-D images. We set different parameter values to mine mVAPs, and the evaluation method follows the same cross-validation procedure as in Experiment 1. In particular, a new type of ARGs are used to represent objects in cluttered indoor RGB images.

\begin{figure}
\centering
\includegraphics[width=\linewidth]{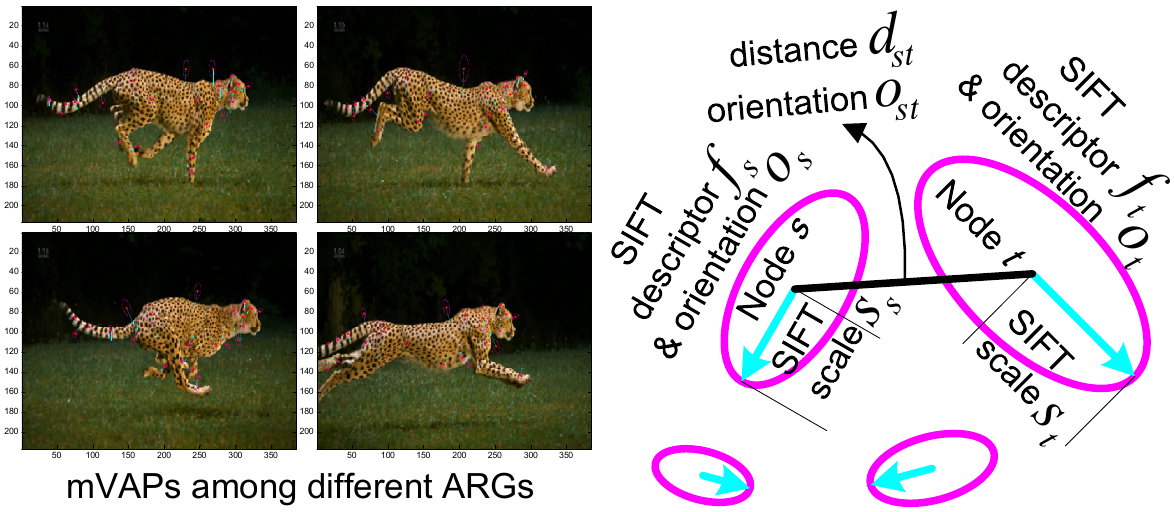}
\caption{Notation for the ARGs that take SIFT points as nodes and are used in Experiments 3 and 5}
\label{fig:ARG_SIFT}
\end{figure}

\textbf{Edge-based ARGs for RGB images:}{\verb| |} The ARGs use edge segments as graph nodes, just like those for RGB-D images. A total of one type of unary attribute ($n^{U}=1$) and three types of pairwise attributes ($n^{P}=3$) are designed. Please see Fig.~\ref{fig:ARG_edge} for notation.

The only unary attribute is defined as the HoG features at the two line terminals, as for RGB-D images. The first pairwise attribute between each pair of nodes $s$ and $t$ represents the 2D angle between their line segments, which is given by ${\mathcal F}_{1}^{st}=\theta^{2D}_{st}$. Then, the angles between the centerline of $(s,t)$ and the two line segments are defined as the second pairwise attribute, denoted by ${\mathcal F}_{2}^{st}=[\theta^{center}_{s},\theta^{center}_{t}]$, where $\theta^{center}_{s}$ represents the angle between the line of $s$ and the centerline. The third pairwise attribute is given by ${\mathcal F}_{3}^{st}=\frac{1}{l_{center}}[l^{2D}_{s},l^{2D}_{t}]$, where $l^{2D}_{s}$ and $l_{center}$ are the lengths of line segment $s$ and the centerline, respectively.

\begin{figure*}
\centering
\includegraphics[width=\linewidth]{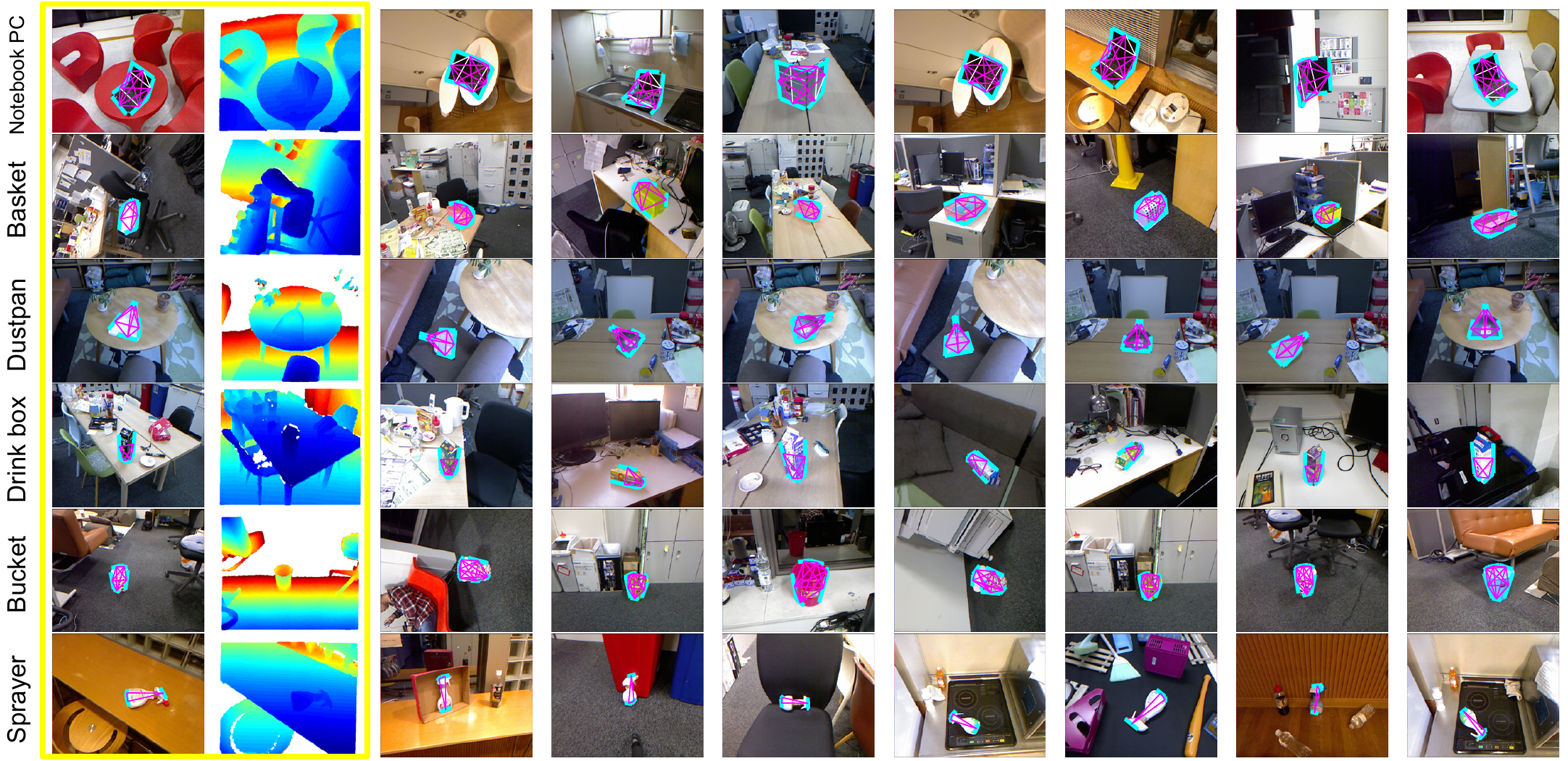}
\caption{Object matching using mVAPs mined from RGB-D images (\emph{i.e.} ARGs taking edges as nodes) in Experiment 1. Magenta/white edges denote directed edges with single/double orientation(s).}
\label{fig:result_RGBD}
\end{figure*}

\subsection{Experiment 3: mining SIFT-based models from web images}

In this experiment, we apply our method to a more general visual data, \emph{i.e.} web images. Ten keywords (``bag,'' ``boot,'' ``camera,'' ``coca cola,'' ``glasses,'' ``hamster,'' ``iphone,'' ``panda,'' ``sailboat,'' and ``spider'') are used to collect internet images for ten categories. For each category, we select the 200 top-ranked images to construct the ARGs. This dataset has been published in \cite{WebImageDataset} and used to test conventional graph mining~\cite{OurCVPR14Graph}. In this dataset, the initial graph templates are also provided.

We apply different values of $\tau$ to test the mining performance. Note that, in ubiquitous web images, we cannot ensure there exists a single pattern that is able to describe target objects in all the images. Therefore, we set two criteria to control the image quality during the mining process. First, we require that during \textbf{Op. 1} in each iteration, more than $60\%$ of the pattern nodes should be matched to the nodes in the positive ARGs, rather than \textit{none}. In other words, if the matched subgraphs in positive ARGs represent less than $60\%$ of the pattern size, these subgraphs will be considered invalid and removed. Second, in each iteration, no more than 20 web images should be used for model mining. We simply select the 20 top-ranked images from those collected, \emph{i.e.} the 20 images (ARGs) with the lowest matching energies.

\textbf{ARGs based on SIFT points:}{\verb| |} We use a new kind of ARG that selects points of interest from the SIFT features in each image as the graph nodes. The SIFT-based ARGs are designed for more general images, especially those without clear edges. We design two local features ($N_{P}=2$) and five pairwise attributes ($N_{Q}=5$) for this ARG type (please see Fig.~\ref{fig:ARG_SIFT} for notation). Let $f_{s}$, $o_{s}$, ${\bf p}_{s}$, and ${\bf s}_{s}$ be the 128-dimensional descriptor, orientation, position, and scale of node $s$'s SIFT feature. We set the unary attributes for each node $s$ as ${\mathcal F}_{1}^{s}=f_{s}$ and ${\mathcal F}_{2}^{s}=o_{s}$, and set six pairwise attributes for each edge $(s,t)$ as ${\mathcal F}_{1}^{st}\!=\!angle({\bf o}_{s},{\bf o}_{t})$, ${\mathcal F}_{2}^{st}\!=\!angle({\bf o}_{s},{\bf p}_{s}\!-\!{\bf p}_{t})$,${\mathcal F}_{3}^{st}\!=\!angle({\bf o}_{t},{\bf p}_{s}\!-\!{\bf p}_{t})$, ${\mathcal F}_{4}^{st}\!=\!{\ln}({\bf s}_{s}/{\bf s}_{t})$, ${\mathcal F}_{5}^{st}\!=\!{\ln}(\sqrt{{\bf s}_{s}^2\!+\!{\bf s}_{t}^2}/\Vert{\bf p}_{s}\!-\!{\bf p}_{t}\Vert)$, ${\mathcal F}_{6}^{st}\!=\!{\bf p}_{s}\!-\!{\bf p}_{t}/\Vert{\bf p}_{s}\!-\!{\bf p}_{t}\Vert$.

\subsection{Experiment 4: mining models from Pascal VOC2007}

In this experiment, we mine the models from one of the most challenging visual datasets, Pascal VOC2007~\cite{pascal-voc-2007}. We apply our method to the Pascal VOC2007 $6\times2$ dataset, which contains \textit{bus}, \textit{motorbike}, \textit{aeroplane}, \textit{horse}, \textit{boat}, and \textit{bicycle} categories with \textit{left} and \textit{right} poses.

Note that, because Pascal VOC images have significant intra-category appearance variations due to occlusions and texture changes, no specific pattern can successfully describe all training images. Therefore, we only mine the pattern corresponding to the top-$10\%$ of images, \emph{i.e.} we determine the $10\%$ of the training images with the lowest matching energies in \textbf{Op.}~1 in each iteration, and use these for the further mining process. Consequently, in Table~\ref{tab:detection_VOC}, we evaluate the pattern's expressive power on the top-$10\%,20\%,...,50\%$, and $100\%$ of the Pascal VOC2007 testing images.

\begin{figure}
\centering
\includegraphics[width=\linewidth]{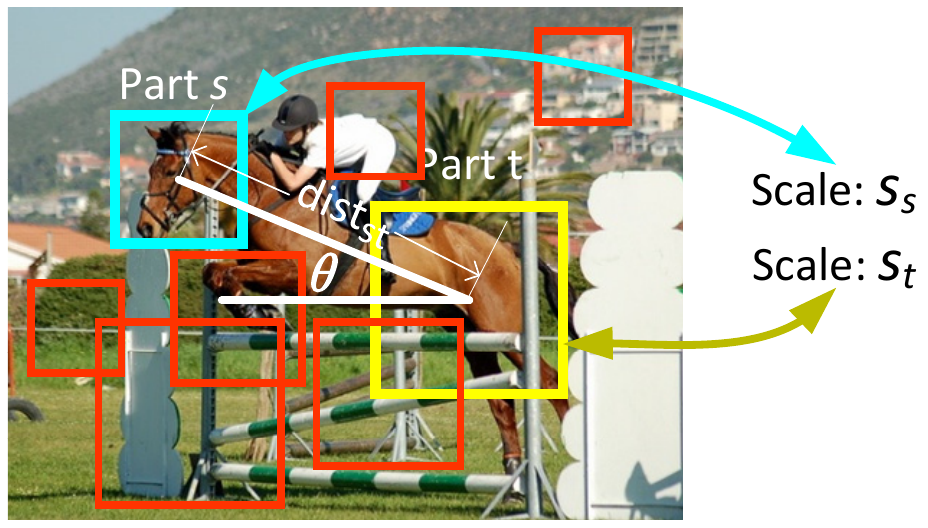}
\caption{Notation for the ARGs in Experiment 4 that take middle-level patches as nodes}
\label{fig:ARG_MidPatch}
\end{figure}

\textbf{ARGs based on middle-level patches:}{\verb| |} We construct ARGs using middle-level patch features~\cite{MiddleLevel} as nodes and taking their HoG features as the only unary attributes. Let $S_{s}$ and $(u_{s},v_{s})$ denote the patch scale and 2D coordinates of node $s$. Three pairwise attributes ($N_{Q}\!=\!3$) are used for each edge $(s,t)$, \emph{i.e.} ${\mathcal F}_{1}^{st}\!=\!\frac{1}{dist_{st}}[u_{s}-u_{t},v_{s}-v_{t}]^{T}$,
${\mathcal F}_{2}^{st}\!=\!\log({S_{s}}/{S_{t}})$,
${\mathcal F}_{3}^{st}\!=\![\log({S_{s}}/{dist_{st}}),\log({S_{t}}/{dist_{st}})]$, where $dist_{st}=\sqrt{(u_{s}-u_{t})^2+(v_{s}-v_{t})^2}$.

\subsection{Experiment 5: mining SIFT-based models from videos}

In this experiment, we collect video sequences for a cheetah, swimming girls, and a frog from the internet. Each frame of these videos is formulated as an ARG. As in Experiment 3, the ARGs are constructed based on SIFT points. Thus, we use our method to mine mVAPs from these videos as the models for deformable objects. The initial graph templates only contain three nodes.

\subsection{Competing methods}
\label{sec:competing}

We comprehensively compare our method with seven competing methods, including three image/graph matching approaches, two methods of unsupervised learning for graph matching, one approach for refining the pattern structure, and one pioneering technique for visual graph mining. All the competing methods are designed considering the same scenario of ``learning models from a number of unlabeled ARGs with a single labeled object.'' To enable a fair comparison, all competing methods are provided with the same initial graph templates, as well as the same sets of training ARGs and testing ARGs.

First, image/graph matching approaches without any learning techniques are taken as the baseline. Generally speaking, there are two typical paradigms for image matching. One is the minimization of matching energy as in this study, and the other is the maximization of matching compatibility, where matching compatibility is usually defined as $\arg\!\max_{{\bf x}}{\mathcal C}({\bf x})\!=\!\!\sum_{s,t}\!{\Large e}^{-P_{s}(x_{s})-P_{t}(x_{t})-Q_{st}(x_{s},x_{t})}$. Thus, three competing methods, \emph{i.e.} \textit{MA},\textit{MS}, and \textit{MT}, are designed to represent these two paradigms. \textit{MA} uses TRW-S~\cite{TRWS} to minimize the matching energy in (\ref{eqn:energy}). \textit{MS} and \textit{MT} maximize the overall matching compatibility ${\mathcal C}({\bf x})$ (proposed in \cite{GraphMatchingCMU_M,LearnMatchingCMU}), in which $P_{s}(\cdot)$ and $Q_{st}(\cdot,\cdot)$ are defined using absolute attribute differences. \textit{MS} and \textit{MT} use spectral techniques~\cite{GraphMatchingCMU_M} and TRW-S~\cite{TRWS}, respectively, to compute $\arg\!\max_{{\bf x}}{\mathcal C}({\bf x})$.

\begin{figure*}
\centering
\includegraphics[width=\linewidth]{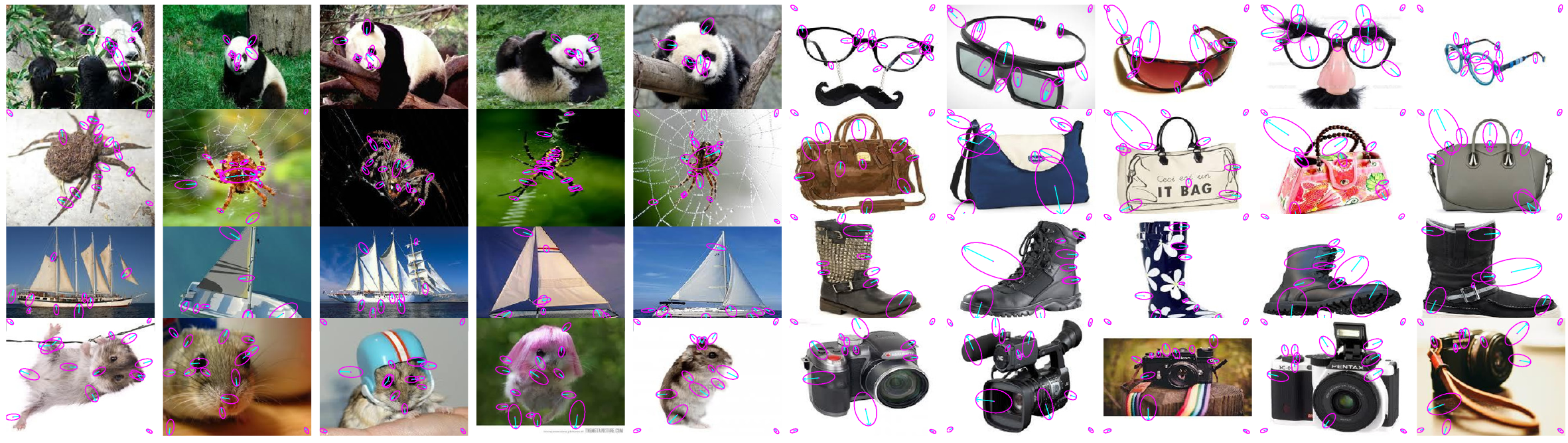}
\caption{Object matching using mVAPs mined from web images (\emph{i.e.} ARGs taking SIFT points as nodes) in Experiment 3. Considering copyright reasons, results for the ``iphone'' and ``coca cola'' categories are not shown.}
\label{fig:result_SIFT}
\end{figure*}

\begin{figure*}
\centering
\includegraphics[width=\linewidth]{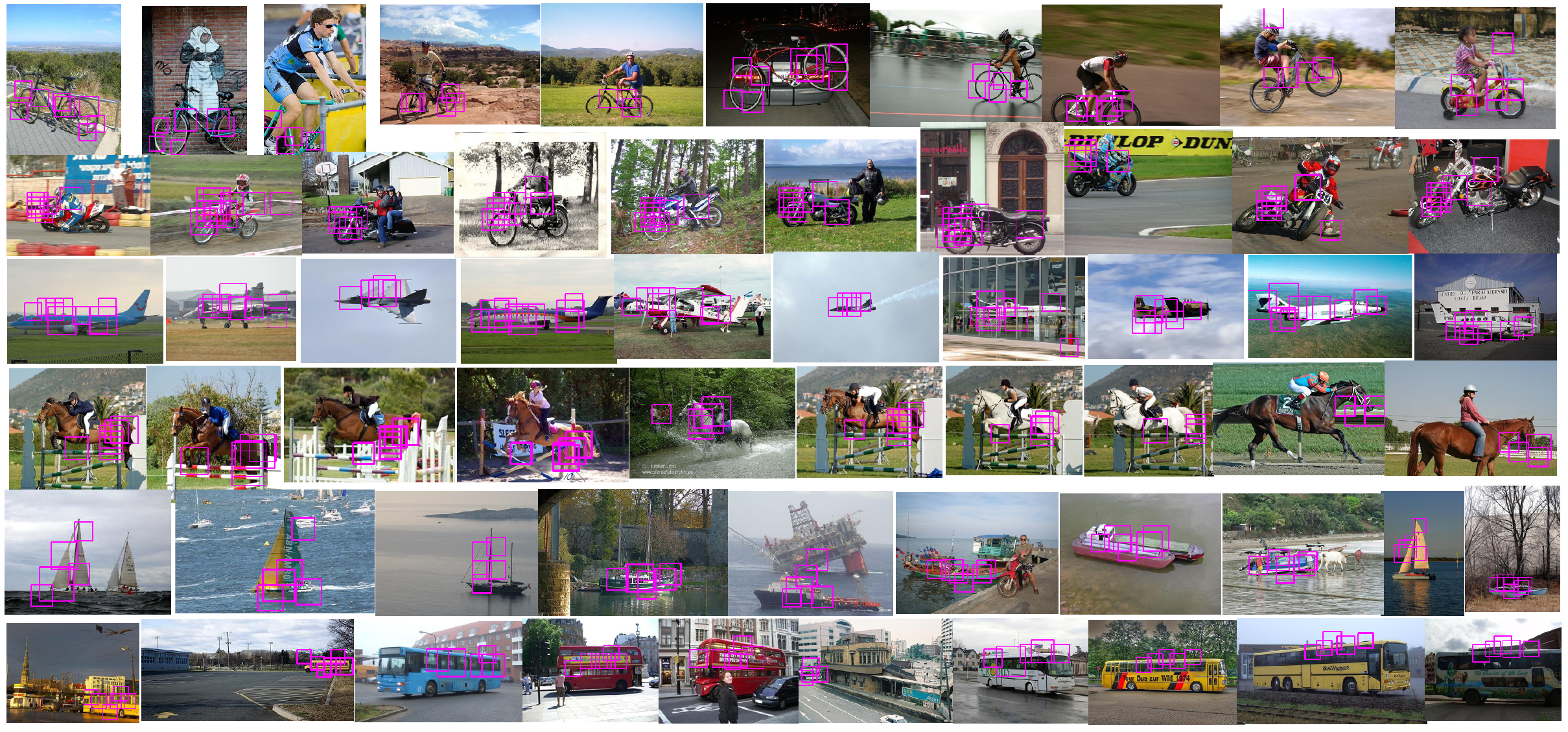}
\caption{Object matching using mVAPs mined from Pascal VOC2007 images (\emph{i.e.} ARGs taking middle-level patches as nodes) in Experiment 4.}
\label{fig:result_VOC}
\end{figure*}

\begin{figure*}
\centering
\includegraphics[width=\linewidth]{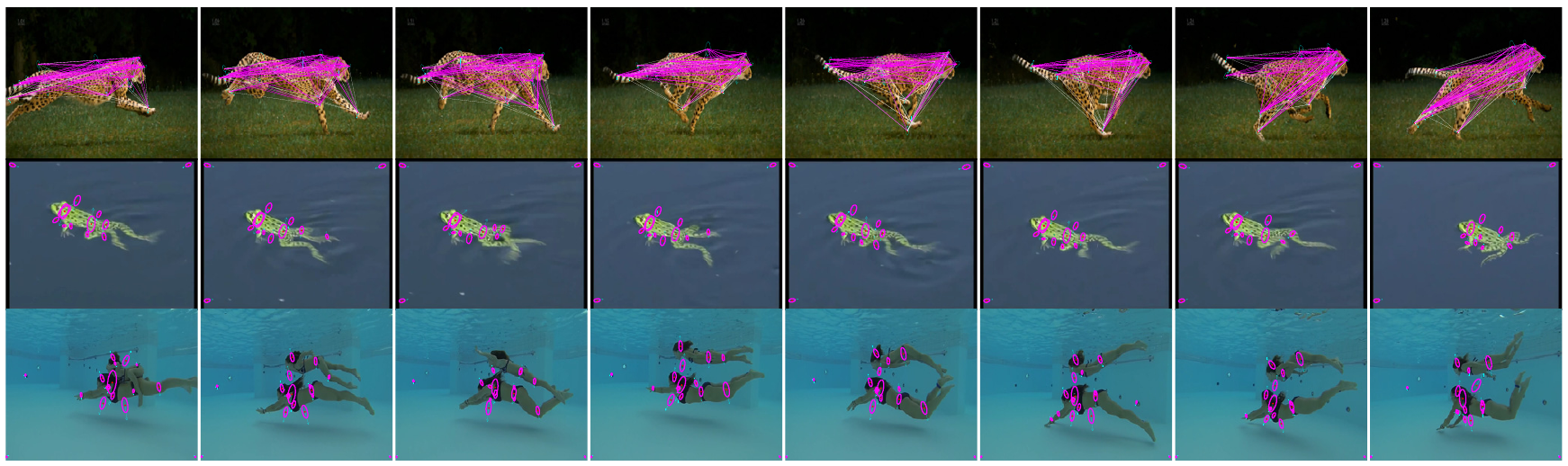}
\caption{Object matching using mVAPs mined from video frames (\emph{i.e.} ARGs taking SIFT points as nodes) in Experiment 5. In the first row, magenta/white edges denote directed edges with single/double orientation(s).}
\label{fig:result_video}
\end{figure*}

\begin{table*}
\caption{Comparison of the APs of patterns mined from web images. The mVAPs are mined by setting $d\!=\!2$.}
\label{tab:detection}
\begin{center}
\resizebox{1.0\hsize}{!}{
\begin{tabular}{|c|c|c|c|c|c|c|c|c|c|c|c|}
\hline
\!\!{\footnotesize AP$\uparrow$}\!\! &\multicolumn{1}{|c|}{\textit{\footnotesize bag}} & \multicolumn{1}{|c|}{\textit{\footnotesize boot}} & \multicolumn{1}{|c|}{\textit{\footnotesize camera}} & \multicolumn{1}{|c|}{\textit{\footnotesize coca cola}} & \multicolumn{1}{|c|}{\textit{\footnotesize glasses}} & \multicolumn{1}{|c|}{\textit{\footnotesize hamster}} & \multicolumn{1}{|c|}{\textit{\footnotesize iphone}} & \multicolumn{1}{|c|}{\textit{\footnotesize panda}} & \multicolumn{1}{|c|}{\textit{\footnotesize sailboat}} & \multicolumn{1}{|c|}{\textit{\footnotesize spider}} & \multicolumn{1}{|c|}{{\small\bf Average}}\\
\cline{1-1}
\!\!$\tau$\!\! & \!\!\!\textit{\footnotesize mSAP,\, Our}\!\!\! & \multicolumn{1}{|c|}{\!\!\!\textit{\footnotesize mSAP,\, Our}\!\!\!} & \multicolumn{1}{|c|}{\!\!\!\textit{\footnotesize mSAP,\, Our}\!\!\!} & \multicolumn{1}{|c|}{\!\!\!\textit{\footnotesize mSAP,\, Our}\!\!\!} & \multicolumn{1}{|c|}{\!\!\!\textit{\footnotesize mSAP,\, Our}\!\!\!} & \multicolumn{1}{|c|}{\!\!\!\textit{\footnotesize mSAP,\, Our}\!\!\!} & \multicolumn{1}{|c|}{\!\!\!\textit{\footnotesize mSAP,\, Our}\!\!\!} & \multicolumn{1}{|c|}{\!\!\!\textit{\footnotesize mSAP,\, Our}\!\!\!} & \multicolumn{1}{|c|}{\!\!\!\textit{\footnotesize mSAP,\, Our}\!\!\!} & \multicolumn{1}{|c|}{\!\!\!\textit{\footnotesize mSAP,\, Our}\!\!\!} &
\multicolumn{1}{|c|}{\!\!\!{\bf mSAP,\, Our}\!\!\!}\\
\cline{2-11}
\!\!{\footnotesize 1/32}\!\!
 & \!\! 72.4 \,{\bf 84.8}\!
 & \!\! 71.4 \,{\bf 93.7}\!
 & \!\! 79.0 \,{\bf 90.1}\!
 & \!\! 65.2 \,{\bf 66.9}\!
 & \!\! {\bf 73.0} \, 69.2\!
 & \!\! 50.7 \,{\bf 84.8}\!
 & \!\! 81.9 \,{\bf 92.9}\!
 & \!\! 90.5 \,{\bf 93.8}\!
 & \!\! {\bf 69.4} \, 68.3\!
 & \!\! 80.1 \,{\bf 94.5}\!
 & \!\! 73.3 \,{\bf 83.9}\!\\
\!\!{\footnotesize 2/32}\!\!
 & \!\! {\bf 90.9} \,84.9\!
 & \!\! {\bf 76.4} \,68.0\!
 & \!\! {\bf 85.5} \,79.1\!
 & \!\! 80.2 \,{\bf 88.0}\!
 & \!\! {\bf 69.6} \,69.3\!
 & \!\! 55.5 \,{\bf 82.4}\!
 & \!\! {\bf 76.7} \,68.4\!
 & \!\! {\bf 94.8} \,94.1\!
 & \!\! {\bf 67.6} \,66.6\!
 & \!\! 57.2 \,{\bf 74.3}\!
 & \!\! 75.4 \,{\bf 77.5}\!\\
\!\!{\footnotesize 3/32}\!\!
 & \!\! 89.6 \,{\bf 92.7}\!
 & \!\! 76.9 \,{\bf 80.0}\!
 & \!\! 83.9 \,{\bf 84.2}\!
 & \!\! 67.6 \,{\bf 82.2}\!
 & \!\! 77.9 \,{\bf 83.9}\!
 & \!\! 45.8 \,{\bf 82.4}\!
 & \!\! {\bf 78.0} \,68.4\!
 & \!\! {\bf 94.9} \,94.3\!
 & \!\! {\bf 67.9} \,66.6\!
 & \!\! 82.2 \,{\bf 86.0}\!
 & \!\! 76.5 \,{\bf 82.0}\!\\
\!\!{\footnotesize 4/32}\!\!
 & \!\! 90.5 \,{\bf 92.5}\!
 & \!\! 80.9 \,{\bf 81.1}\!
 & \!\! 86.9 \,{\bf 88.3}\!
 & \!\! {\bf 73.6} \,66.9\!
 & \!\! 84.6 \,{\bf 86.1}\!
 & \!\! {\bf 55.5} \,54.2\!
 & \!\! {\bf 78.3} \,67.6\!
 & \!\! {\bf 95.8} \,95.7\!
 & \!\! {\bf 63.4} \,49.1\!
 & \!\! 82.1 \,{\bf 87.7}\!
 & \!\! {\bf 79.2} \,76.9\!\\
\!\!{\footnotesize 5/32}\!\!
 & \!\! 90.5 \,{\bf 92.4}\!
 & \!\! 76.8 \,{\bf 82.2}\!
 & \!\! 87.0 \,{\bf 92.0}\!
 & \!\! 78.1 \,{\bf 81.4}\!
 & \!\! 82.0 \,{\bf 95.2}\!
 & \!\! {\bf 53.9} \,52.5\!
 & \!\! 73.7 \,{\bf 79.0}\!
 & \!\! 94.5 \,{\bf 97.7}\!
 & \!\! 62.5 \,{\bf 68.3}\!
 & \!\! 82.7 \,{\bf 88.4}\!
 & \!\! 78.2 \,{\bf 82.9}\!\\
\!\!{\footnotesize 6/32}\!\!
 & \!\! 88.0 \,{\bf 91.6}\!
 & \!\! 78.5 \,{\bf 79.0}\!
 & \!\! 88.6 \,{\bf 92.3}\!
 & \!\! 73.5 \,{\bf 82.7}\!
 & \!\! 82.3 \,{\bf 88.1}\!
 & \!\! 54.6 \,{\bf 56.8}\!
 & \!\! 76.4 \,{\bf 82.9}\!
 & \!\! 93.5 \,{\bf 96.7}\!
 & \!\! 63.5 \,{\bf 69.8}\!
 & \!\! 83.1 \,{\bf 90.6}\!
 & \!\! 78.2 \,{\bf 83.0}\!\\
\hline
\end{tabular}}
\end{center}
\end{table*}

\begin{figure*}
\centering
\includegraphics[width=\linewidth]{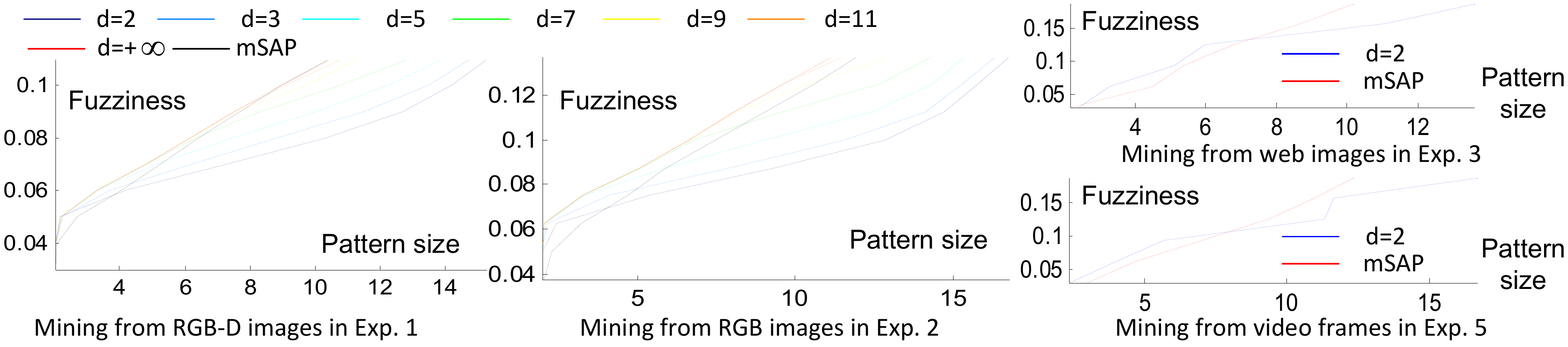}
\caption{Comparison of pattern fuzziness}
\label{fig:graphsize}
\end{figure*}

Second, we compare the proposed method with unsupervised approaches for learning graph matching, \emph{i.e.} those that do not require people to label matching assignments. These techniques mainly learn matching parameters, but do not consider the learning of pattern structure, to achieve good matches between a graph template and a set of ARGs. Leordeanu \emph{et al.}~\cite{LearnMatchingCMU} proposed the benchmark of unsupervised learning for graph matching. We design two competing methods to represent this technique, \emph{i.e.} \textit{LS} and \textit{LT}. These two methods use \cite{GraphMatchingCMU_M} and \cite{TRWS}, respectively, to solve the graph-matching optimization $\arg\!\max_{{\bf x}}{\mathcal C}({\bf x})$. They iteratively train the attribute weights, \emph{i.e.} $w^{P}_{i}$ and $w^{Q}_{j}$ in the matching compatibility ${\mathcal C}({\bf x})$ mentioned above.

Third, we compare our method with the structural refinement method~\cite{OurICCV13}, denoted by \textit{SR}. Actually, we can regard this method as the boundary between graph mining and unsupervised learning for graph matching. \textit{SR} refines the pattern's structure by simultaneously deleting ``bad'' nodes, training matching parameters, and estimating attributes, but the key for graph mining, \emph{i.e.} the discovery of new pattern nodes, is not involved. Note that when we apply different values of $\tau$, our method will produce mVAPs with different graph sizes. Therefore, to enable a fair comparison, we require \textit{SR} to modify the initial graph template to a pattern with the same size as the mVAP that is mined for a given $\tau$.

Finally, we compare the proposed method with the only pioneering method that mines subgraph patterns, namely, maximal-size soft attributed patterns (\textit{mSAP}), from visual data~\cite{OurCVPR14Graph}. This method considers both the expressive power of graph mining and the challenges of visual mining. However, \cite{OurCVPR14Graph} requires the parameters ${\bf w}$, $P_{none}$ and $Q_{none}$ to be manually set. Thus, to enable a fair comparison, we apply the training of $P_{none}$ and $Q_{none}$ proposed in this study to \cite{OurCVPR14Graph}, and initialize ${\bf w}$ as in our method.

\subsection{Evaluation metrics, results, and analysis}

\begin{figure}
\centering
\includegraphics[width=\linewidth]{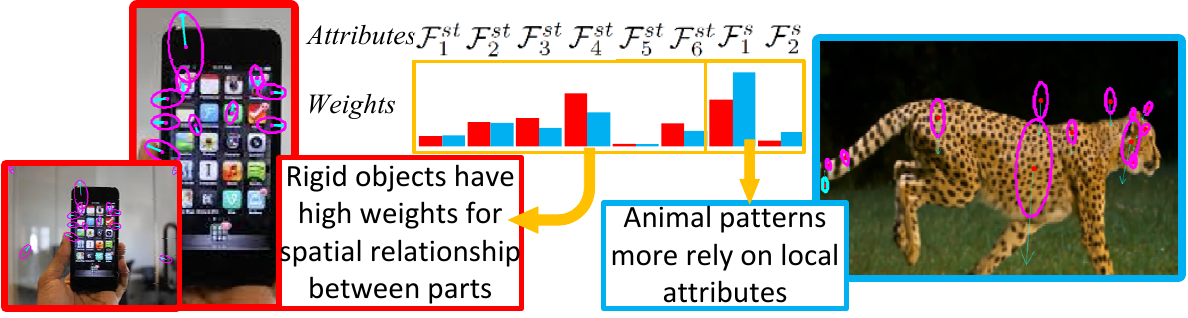}
\caption{Attribute weights mined for a rigid pattern in Experiment 3 and a animal pattern in Experiment 5}
\label{fig:attributeCompare}
\end{figure}

Figs.~\ref{fig:result_RGBD}, \ref{fig:result_SIFT}, \ref{fig:result_VOC}, and \ref{fig:result_video} show the objects corresponding to the patterns mined in Experiments 1, 3, 4, and 5, respectively. Fig.~\ref{fig:attributeCompare} compares the attribute weights that are trained for a static iPhone in Experiment 3 and a running cheetah in Experiment 5. Static objects usually have higher weights on the pairwise relationship between object parts, whereas patterns for dynamic animals focus more on local attributes.

\textbf{Pattern fuzziness: }{\verb| |} In the experiments, we control the pattern fuzziness $\tau$ to obtain patterns with different sizes. In Fig.~\ref{fig:graphsize}, we show the relationship between the pattern fuzziness and the average size of the patterns mined under this fuzziness. Because \cite{OurCVPR14Graph} defined the fuzziness of the mSAPs in a similar way, we can compare the pattern fuzziness between the mSAPs and the mVAPs mined using different values of $d$. As in our method, \cite{OurCVPR14Graph} can mine mSAPs with different sizes given different fuzziness settings. Only patterns of similar size can be fairly compared.

Fig.~\ref{fig:graphsize} shows that the mVAPs produced by our method are less fuzzy than the mSAPs mined by \cite{OurCVPR14Graph}. More sparse mVAPs (with smaller values of $d$) usually have less fuzziness. Theoretically, if we ignore the training of matching parameters, the mining of mSAPs is equivalent to our mining of mVAPs when $d\!=\!+\infty$. Thus, the curves of mSAPs and mVAPs with $d\!=\!+\infty$ exhibit some similarities.

\textbf{Energy gap between positive and negative matches: }{\verb| |} We can use the mined patterns to match the target objects in previously unseen positive ARGs and negative ARGs. Therefore, we can regard the ratio of the average energy of the positive matches to that of the negative matches as a metric to evaluate the distinguishing ability of the pattern. This is similar to the ``eigengap'' for evaluating the spectral graph matching~\cite{GraphMatchingCMU_M,LearnMatchingCMU}. Patterns with low positive-negative energy ratios usually have a stable matching performance.

\begin{figure}
\centering
\includegraphics[width=\linewidth]{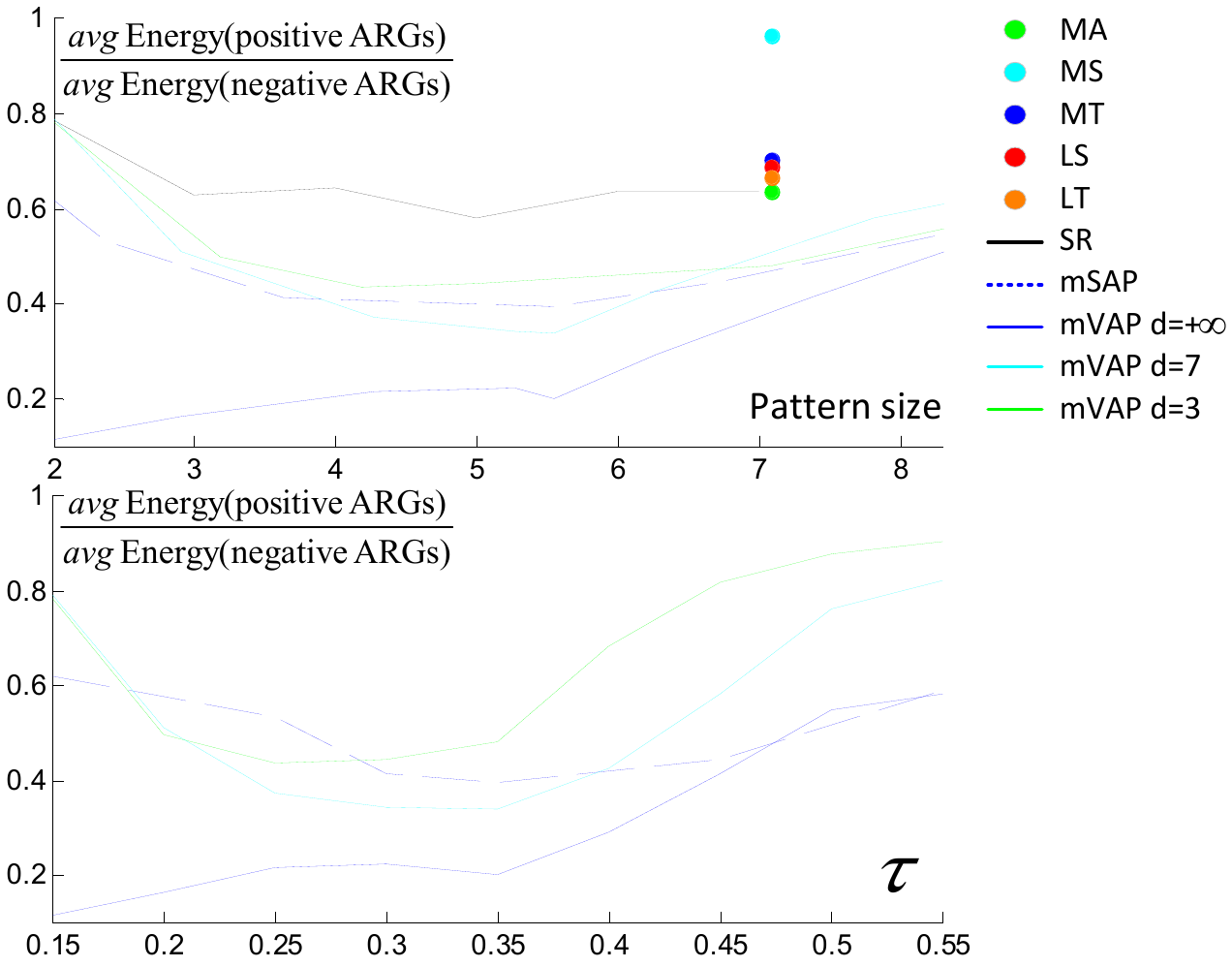}
\caption{Ratio of the energies of positive matches to those of negative matches. (top) We compare the energy ratios between different competing methods. (bottom) We show the ratio changes along with $\tau$.}
\label{fig:ratio}
\end{figure}

In Experiment 1, we produce different sets of \textit{notebook PC} patterns with different sizes by applying different values of $\tau$ via a series of cross-validations. Fig.~\ref{fig:ratio}(top) shows how the average energy ratio changes with the average pattern size among different sets of \textit{notebook PC} patterns. Fig.~\ref{fig:ratio}(bottom) illustrates the performance using different values of $\tau$. The mVAPs exhibit lower energy ratios than the other competing methods. We compare the mSAPs produced by \cite{OurCVPR14Graph} (blue dashed lines) with the mVAPs mined with $d=+\infty$ (blue solid lines), because both of these patterns are complete graphs. Our mVAPs perform better than the mSAPs.

\textbf{Average precision: }{\verb| |} We now test the object detection performance of the mined pattern. As mentioned above, we match the patterns to a set of previously unseen positive and negative ARGs. We use the simplest way of identifying the true and false detections: Given the matching assignments ${\bf\hat{x}}^{k}$ for each ARG $k$ and a threshold, if the value of $[{\mathcal E}_{s}^{k}-\zeta\sum_{s}\delta(\hat{x}_{s}^{k})/\vert{V}\vert]$ (here, $\zeta\!=\!10$) is greater than the threshold, we consider this to be a true detection; otherwise, it is a false detection. Hence, we can draw a precision-recall curve of object detection by choosing different thresholds. The average precision (AP) of the precision-recall curve is used as a metric to evaluate the graph matching performance.

\begin{table}
\caption{Comparison of average APs of the twelve patterns mined from the Pascal VOC2007 $6\times2$ dataset. The mVAPs are extracted by setting $d\!=\!10, \tau\!=\!1.0$.}
\label{tab:detection_VOC}
\begin{center}
\resizebox{1.0\hsize}{!}{
\begin{tabular}{|c|cccccc|}
\hline
testing & top & top & top & top & top & top\\
images & $10\%$ & $20\%$ & $30\%$ & $40\%$ & $50\%$ & $100\%$\\
\hline
\textit{\footnotesize mSAP} &
68.9 & 46.0 & 33.3 & 26.0 & 21.5 & 11.8\\
\textit{\footnotesize Our} &
{\bf 80.9} & {\bf 50.5} & {\bf 37.7} & {\bf 30.0} & {\bf 24.7} & {\bf 13.5} \\
\hline
\end{tabular}}
\end{center}
\end{table}

\begin{figure}
\centering
\includegraphics[width=\linewidth]{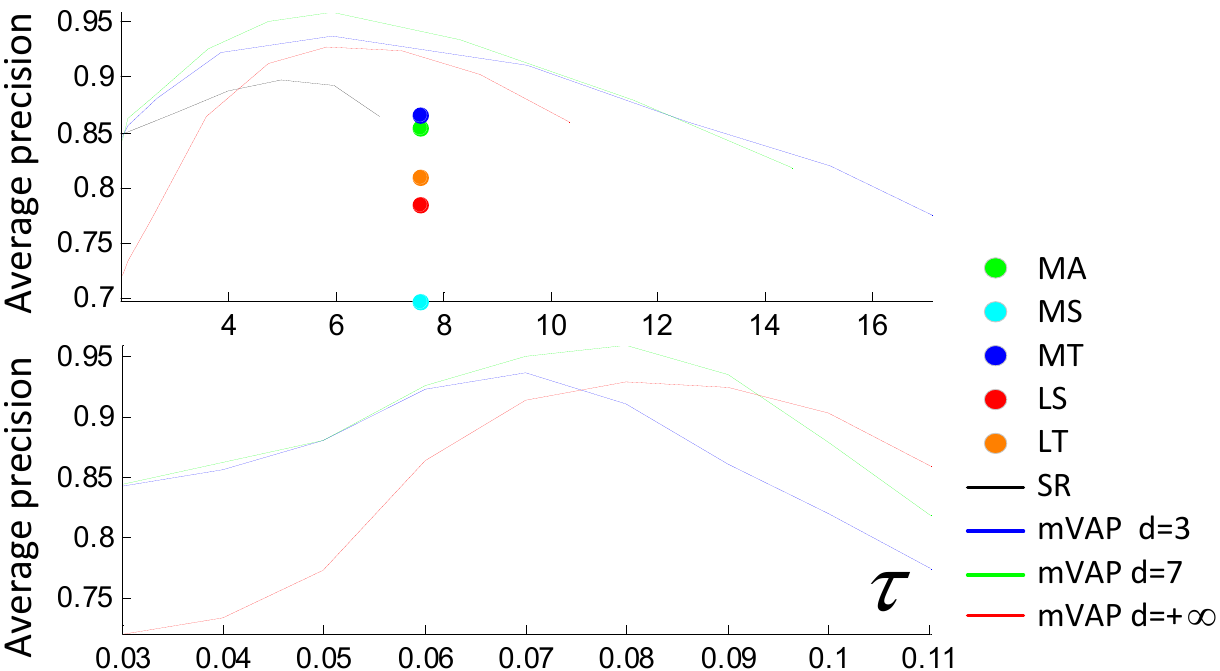}
\caption{APs of object detection using the mined patterns. (top) We compare the APs between different competing methods. (bottom) We show the AP changes along with $\tau$.}
\label{fig:AP}
\end{figure}

Table~\ref{tab:detection} and Table~\ref{tab:detection_VOC} shows that our mVAPs are more distinguishing than mSAPs. In Experiment 1, we use different values of $\tau$ to produce different sets of patterns with different average sizes. Fig.~\ref{fig:AP}(top) shows how the average AP of the patterns changes with their size for different pattern sets. In Fig.~\ref{fig:AP}(bottom), we plot the curves that illustrate the relationship between the average AP of the mVAPs and the value of $\tau$. In Fig.~\ref{fig:AP}(top), we compare our method with a total of six competing methods. It can be seen that our method exhibits superior performance to the other approaches. Fig.~\ref{fig:AP} gives a clear comparison between the mVAPs mined with different values of $d$. Compared to mining complete subgraph patterns (\emph{i.e.} setting $d=+\infty$), the combination of mining pattern linkages increases the detection accuracy. This figure also demonstrates that the mining process may drift if the pattern is modified to contain too few or too many nodes. Small pattern fuzziness ($\tau$) produces mVAPs with a small number of nodes, which lack enough information for reliable object detection, while large fuzziness makes the pattern contain unreliable nodes and decreases the performance.

\begin{figure}
\centering
\includegraphics[width=\linewidth]{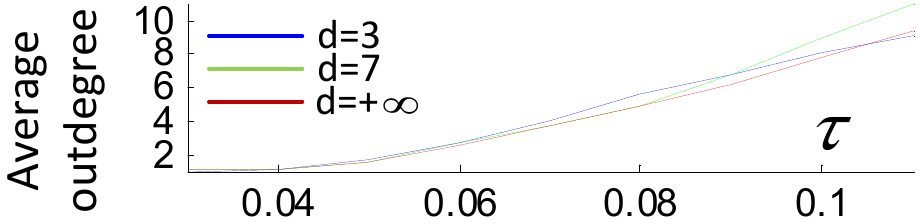}
\caption{Average number of latent linkages mined for each node. The linkage mining process is insensitive to $d$.}
\label{fig:density}
\end{figure}

\textbf{Latent pattern density: }{\verb| |} Parameter $d$ in Definition~\ref{def:VAP} describes the minimum number of outgoing edges and is not a strong control of pattern density. Given a pattern fuzziness $\tau$, the latent pattern density is automatically mined. Fig.~\ref{fig:density} shows the average number of outgoing edges for each node mined using different parameters in Experiment 1. It demonstrates that the number of outgoing edges for each node is mainly controlled by the pattern's fuzziness level $\tau$. If we set a large value for $d$, \emph{e.g.} $d\!=\!+\infty$, then Fig.~\ref{fig:graphsize} indicates that we would obtain small patterns, which in return limits the pattern's edge number.

\begin{figure}
\centering
\includegraphics[width=\linewidth]{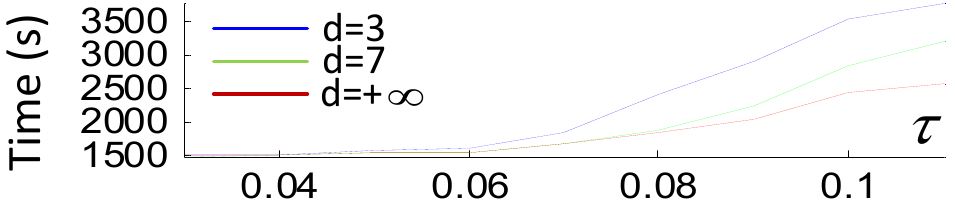}
\caption{Time costs in Experiment 1}
\label{fig:time}
\end{figure}

\textbf{Computation time of learning/mining: }{\verb| |} Fig.~\ref{fig:time} shows the average computational cost of mining each category model in Experiment 1. We implemented the algorithm using Matlab, and computed the running time using eight hyper threads on a machine with an Intel Xeon CPU X5560 @2.80GHz. Larger values of $\tau$ usually produce larger patterns, and require more computational time.

We now briefly analyze the computational cost of the competing methods. The main computational task of these approaches is the energy minimization (or compatibility maximization) of graph matching during the learning/mining process. First, image/graph matching approaches, such as \textit{MA}, \textit{MS}, and \textit{MT}, do not apply any learning techniques.

Second, unsupervised methods for learning graph matching (\textit{LS} and \textit{LT}) use $M$ iterations to learn the matching parameters without changing the size of the graph template. In each iteration, they match the graph template to all $N^{+}$ positive ARGs. We use $n^{0}$ and $n_{k}^{+}$ to denote the node number of the initial graph template and the node number of the $k$-th positive ARG ${\mathcal G}_{k}^{+}$, respectively. The matching to ${\mathcal G}_{k}^{+}$ can be computed as a QAP that assigns each of the $n^{0}$ template nodes to one of $n_{k}^{+}$ labels\footnote{Here, matching choices of \textit{none} are ignored.}. Note that many techniques can be applied to the QAP of matching optimization, and each of them has its own accuracy and computational cost (please refer to \cite{CompareMRFOptimization} for details). Thus, we simply use $c(n^{0}\rightarrow n_{k}^{+})$ to denote the computational cost of this QAP. Larger values of $n^{0}$ and $n_{k}^{+}$ will result in higher cost. Therefore, the computational cost of unsupervised learning can be formulated as $M\sum_{k\!=\!1}^{N^{+}}c(n^{0}\rightarrow n_{k}^{+})$.

Third, for the mining of mSAPs~\cite{OurCVPR14Graph}, let $n^{1},n^{2},...,n^{M}$ denote the node number of the pattern after $1,2,...,M$ iterations. Because this method only uses positive ARGs for training, the computational cost of graph matching is $\sum_{m\!=\!0}^{M-1}\sum_{k\!=\!1}^{N^{+}}c(n^{m}\rightarrow n_{k}^{+})$. In addition, the computational cost of node discovery in each iteration $m$ can be formulated as a QAP that assigns each of the $N^{+}$ positive ARGs with one of $\max_{k}\{n_{k}^{+}-n^{m}\}$ labels (\emph{i.e.} determining the node corresponding to the new node $y$ in each positive ARG). Hence, its computational cost is $c(N^{+}\rightarrow\max_{k}\{n_{k}^{+}-n^{m}\})$. Thus, we can summarize the overall computational cost as $\sum_{m\!=\!0}^{M-1}[c(N^{+}\rightarrow\max_{k}\{n_{k}^{+}-n^{m}\})\!+\!\sum_{k\!=\!1}^{N^{+}}c(n^{m}\rightarrow n_{k}^{+})]$.

Fourth, let us focus on \textit{SR}. Because \textit{SR} only deletes bad nodes without adding new nodes, we assume that the pattern size is $\min\{n^{1},n^{0}\},...,\min\{n^{M},n^{0}\}$ after $1,...,M$ iterations to enable a fair comparison. Consequently, its time cost can be written as $\sum_{m\!=\!0}^{M-1}\sum_{k\!=\!1}^{N^{+}}c(\min\{n^{m},n^{0}\}\rightarrow n_{k}^{+})$.

Finally, let us analyze the proposed method. In addition to the positive matches, our method matches the current pattern to all $N^{-}$ negative ARGs in each iteration. Moreover, unlike the mining of mSAPs~\cite{OurCVPR14Graph}, the QAP for node discovery must be solved twice to detect each new pattern node. Therefore, the overall computational cost of graph mining is $\sum_{m\!=\!0}^{M-1}[2c(N^{+}\rightarrow\max_{k}\{n_{k}^{+}-n^{m}\})\!+\!\sum_{k\!=\!1}^{N^{+}}c(n^{m}\rightarrow n_{k}^{+})\!+\!\sum_{l\!=\!1}^{N^{-}}c(n^{m}\rightarrow n_{l}^{-})]$, where $n_{l}^{-}$ denotes the node number of the negative ARG ${\mathcal G}_{l}^{-}$. In summary, compared to other methods of pattern learning/mining, our method requires additional computation for the matching to negative ARGs. The additional computation cost is linear with respect to the number of negative ARGs. Besides, the mining of node linkages doubles the computational cost of node discovery. Nevertheless, the proposed method has the same order of time complexity as \cite{OurCVPR14Graph}.


\section{Conclusions and discussions}
\label{sec:conclude}

In this paper, we have extended the scope of mining maximal frequent subgraphs to general visual data, which presents great challenges to both fields of graph mining and computer vision. Considering the challenges in the data collected form real-world situations, we reformulate the concept of subgraph patterns to encode the latent and fuzzy visual knowledge in a general form. The generality of our algorithm has been tested on four kinds of ARGs in five experiments. We have proposed an approximate but practical method to overcome great data fuzziness and comprehensively discover pattern nodes, mine latent node connections and hidden significant attributes, and optimize attribute values and parameters. Compared to the competing methods, our comprehensive pattern mining can better deal with hidden visual uncertainties.

This study aimed to propose a general theoretical solution to graph mining, rather than a sophisticated approach for a specific visual task. In real applications, we can achieve a non-parametric mining process by selecting $\tau$ and $d$ that maximize the AP of pattern-based detections on training images.

In addition, we can add some task-specific techniques to improve the performance. For example, we could add the learning of root templates, extract multiple model components for the category, and apply the non-linear SVM, to compete with supervised
deformable part models~\cite{SupervisedDPM}. We could use segmentation techniques to produce $G^{0}$, thus achieving a fully unsupervised system, or design a hierarchical and-or structure for the pattern to achieve more robustness to object detection. However, in this paper, we simply apply our method to basic ARGs to clarify the method.

{\small
\bibliographystyle{ieee}
\bibliography{TheBib}
}

\end{document}